\pdfoutput=1

\documentclass[11pt]{article}

\usepackage[]{acl}

\usepackage{times}
\usepackage{latexsym}

\usepackage[T1]{fontenc}

\usepackage[utf8]{inputenc}

\usepackage{microtype}

\usepackage{inconsolata}

\usepackage{graphicx}
\usepackage{subcaption}
\usepackage{booktabs}
\usepackage{amsmath}
\usepackage{amsfonts}
\usepackage{algorithm}
\usepackage{algpseudocode}
\usepackage{mathrsfs}
\usepackage{multirow}
\usepackage{caption}
\usepackage{xspace}
\usepackage{float}

\newcounter{notecounter}

\newcommand{\bxbloom}{BX$_\text{BLOOM}$\xspace}
\newcommand{\bxllama}{BX$_\text{LLaMA}$\xspace}
\newcommand{\sftbloom}{SFT$_\text{BLOOM}$\xspace}
\newcommand{\sftllama}{SFT$_\text{LLaMA}$\xspace}
\newcommand{\md}{xLLMs-100\xspace}

\usepackage{soul,xcolor}
\setstcolor{red}

\title{LLMs Beyond English: Scaling the Multilingual Capability of LLMs with Cross-Lingual Feedback}

\author{Wen Lai$^{1,2,3}$\thanks{\hspace{2mm}Work done during internship at Bosch AI.}, Mohsen Mesgar$^{4}$, Alexander Fraser$^{1,3}$\\\\
	$^1$School of Computation, Information and Technology, TUM, Germany \\
        $^2$Center for Information and Language Processing, LMU Munich, Germany \\
        $^3$Munich Center for Machine Learning, Germany \\
        $^4$Bosch Center for Artificial Intelligence, Renningen, Germany\\
        {\tt lavine@cis.lmu.de mohsen.mesgar@bosch.com alexander.fraser@tum.de}
        }

\begin{document}
\maketitle
\begin{abstract}
To democratize large language models (LLMs) to most natural languages, it is imperative to make these models capable of \emph{understanding} and \emph{generating} texts in many languages, in particular low-resource ones. 
While recent multilingual LLMs demonstrate remarkable performance in such capabilities, these LLMs still support a limited number of human languages due to the lack of training data for low-resource languages.
Moreover, these LLMs are not yet aligned with human preference for downstream tasks, which is crucial for the success of LLMs in English.
In this paper, we introduce xLLaMA-100 and xBLOOM-100 (collectively \textbf{\md}), which scale the multilingual capabilities of LLaMA and BLOOM to 100 languages.
To do so, we construct two datasets: a multilingual instruction dataset including 100 languages, which represents the largest language coverage to date, and a cross-lingual human feedback dataset encompassing 30 languages.
We perform multilingual instruction tuning on the constructed instruction data and further align the LLMs with human feedback using the DPO algorithm on our cross-lingual human feedback dataset.
We evaluate the multilingual understanding and generating capabilities of \md on five multilingual benchmarks. 
Experimental results show that \md consistently outperforms its peers across the benchmarks by considerable margins, defining a new state-of-the-art multilingual LLM that supports 100 languages\footnote{The code, datasets, and models are publicly available at \url{https://github.com/boschresearch/ACL24-MLLM}.}.
\end{abstract}

\section{Introduction}
\label{sec:intro}
Despite the impressive improvements have been made recently, the majority of large language models (LLMs), such as LLaMA-2~\cite{touvron2023llama}, LLaMA-3~\cite{llama3modelcard} and BLOOM~\cite{workshop2022bloom}, are  predominantly trained on English texts and support only a limited number of non-English languages.
For instance,  while LLaMA-2 and BLOOM support 25 and 46 languages, respectively, their performance varies significantly across different languages.
However, there are currently more than 7,000 languages spoken in the world\footnote{\url{https://www.ethnologue.com}} and only a few of them are used for training LLMs.
Scaling LLMs' multilingual capabilities is challenging due to the scarcity of multilingual instruction data available for fine-tuning, particularly for low-resource languages.

The primary advantage of LLMs lies in their ability to learn task execution, in particular mapping an input text to an output text, through a textual instruction.
A task instruction, input, and output text are assumed to be in the same language.
However, these elements can be in different languages for many downstream tasks, which is the most intuitive manifestation of the multilingual capabilities of LLMs.
We consider two types of multilingual capability in LLMs.
First, when the instructions for LLMs are expressed in different languages, LLMs should understand these instructions and generate a correct output. 
We refer to this feature of LLMs as the \textit{understanding capability}.
Second, LLMs should be able to generate the correct response in the target language and perform consistently well on (almost) all languages when a fixed language (e.g., English) is used as the instruction language.
We name this capability of LLMs the \textit{generating capability}.

We categorize the previous work on scaling the multilingual capability of LLMs in two groups.
The first group includes approaches that continue training the LLMs using as much of the training corpora as possible. 
The training corpora are either multilingual parallel data for machine translation tasks~\cite{yang2023bigtrans,zhu2023extrapolating}, or multilingual instruction data for instruction tuning~\cite{ustun2024aya, groeneveld2024olmo, luo2023yayi, li2023bactrian, lai-etal-2023-okapi}.
The second group includes approaches that align non-English instructions with English instructions through cross-lingual prompting in the inference stage~\cite{huang-etal-2023-languages, etxaniz2023multilingual}.
While achieving impressive performance, both groups suffer from major problems.
First, they only support a small number of languages, and most of the world's languages are still being left behind.
Second, they use large corpora for supervised fine tuning (SFT), neglecting the exploration of alignment to human preferences.

To address the aforementioned issues, we aim to scale the two multilingual capabilities of LLMs at the same time. 
To improve the understanding capability of LLMs, we construct a multilingual instruction dataset with 100 languages by translating instructions from Alpaca~\cite{alpaca} via ChatGPT\footnote{\url{https://chat.openai.com}} and Google Translate API\footnote{\url{https://translate.google.com}}. 
We fine-tune LLMs on our constructed multilingual dataset using parameter efficient fine tuning (PEFT;~\citealp{hu2021lora}).
The languages of instruction and output are always the same language in the existing human feedback dataset~\cite{alpaca}, which limits the generating capability of LLMs.
To enhance the generating capability, we construct a cross-lingual human feedback data (i.e., instruction and output are different languages) covering 30 languages\footnote{We include 30 languages for that ChatGPT provides high quality feedback.}. 
We then further align the LLMs with human feedback using the DPO algorithm~\cite{rafailov2023direct}.
Finally, we obtain an LLM that supports 100 languages which has the capability to understand the instructions of 100 languages and supports output in 100 languages.

We conduct a comprehensive evaluation to verify the effectiveness of \md on five diverse multilingual benchmarks, covering understanding (PAWS-X;~\citealp{yang-etal-2019-paws}), reasoning (XCOPA;~\citealp{ponti-etal-2020-xcopa}), generation (XL-Sum;~\citealp{hasan-etal-2021-xl} and FLORES-101;~\citealp{goyal-etal-2022-flores}) and expert-written (Self-Instruct;~\citealp{wang-etal-2023-self-instruct}) tasks in the zero-shot setting.
Each benchmark includes multilingual evaluation data covering both high-resource and low-resource languages.
The experimental results clearly demonstrate that \md significantly enhances both the understanding and generating capabilities of LLMs simultaneously across all benchmarks.
Furthermore, our extensive analysis experiments reveal that \md not only mitigate the off-target problem (i.e., LLMs generate the text into an incorrect language~\cite{zhang-etal-2020-improving}) but also enhance language democratization (i.e., democratization degree of tasks between languages~\cite{huang-etal-2023-languages})
compared with strong LLMs.

In summary, we make the following contributions:
\textbf{(i)} We construct two datasets, one of which contains a multilingual instructions in 100 languages, and the other one contains cross-lingual human preferences in 30 languages. 
\textbf{(ii)} We evaluate the multilingual capabilities of LLMs in two dimensions: understanding and generating capability. Unlike previous studies that assess these capabilities in isolation, we urge the community to consider both capabilities when evaluating the multilingual performance of LLMs.
\textbf{(iii)} We scale the multilingual capabilities of LLMs to perform well across 100 languages.
\section{Related Work}
\label{sec:rel_work}
\paragraph{Multilingual Capabilities of LLMs.}
Recent studies have applied LLMs to various NLP tasks in a multilingual setting~\cite{ustun2024aya,groeneveld2024olmo,lai-etal-2023-okapi, weissweiler-etal-2023-counting}.    
In general, two kinds of corpora have been used to improve the multilingual capabilities of LLMs: multilingual parallel corpora (BigTranslate;~\citealp{yang2023bigtrans}) and multilingual instruction datasets (Bactrian-X;~\citealp{li2023bactrian} and xP3;~\citealp{muennighoff-etal-2023-crosslingual}).
\citet{yang2023bigtrans} continued training LLaMA~\cite{touvron2023llama} using a multilingual parallel corpus that covers 102 languages and achieved good results on a multilingual machine translation task. 
However, the results from our initial experiments indicate that the performance of this model on non-machine translation tasks is not as good as its performance on the evaluated machine translation task (see Section~\ref{sec:diff_data} for more details).
\citet{li2023bactrian} finetunes LLaMA-2 using a multilingual instruction dataset including 52 languages, and achieved good performance on several multilingual NLP tasks.
In contrast, we construct a multilingual instruction dataset including 100 languages. 
To the best of our knowledge, our dataset is the multilingual instruction dataset with the largest language coverage to date.
  
\paragraph{Aligning LLMs with Human Feedback.}
A prominent method for LLM training is reinforcement learning from human feedback (RLHF;~\citealp{ouyang2022training}), which learns from human feedback instead of relying on a pre-defined reward function.
Despite its popularity, there are significant flaws. 
Collecting human preferences is time consuming.
The instability of the RLHF method during training also poses a significant challenge in learning the optimal reward function from human preference data~\cite{schulman2017proximal}.
Recently, the DPO algorithm~\cite{rafailov2023direct} has demonstrated that this challenge can be addressed by fine-tuning LLMs to align with human preferences using a supervised learning regime, without the requirement of explicit reward modeling or reinforcement learning.
DPO has only been applied to monolingual human feedback data (i.e., instructions, inputs, and outputs are in the same language).
While such fine-tuned LLMs are beneficial for monolingual tasks, they have not been shown to generalize to tasks that require text understanding and generation in various languages.
In this paper, we construct a large-scale cross-lingual human preference dataset covering 30 languages and show promising results in multiple NLP tasks.
\section{Method}
\label{sec:method}
To scale the multilingual capabilities of LLMs, we construct two datasets: a multilingual instruction dataset (Section~\ref{sec:method_mid}) and a cross-lingual feedback dataset (Section~\ref{sec:method_clfd}).
Then, we evaluate the quality of the translated instructions and generated responses on our constructed datasets (Section~\ref{sec:eval_dataset}).
Finally, we introduce the training process of our multilingual instruction fine-tuning (Section~\ref{sec:method_training}).
Data statistics and analyses can be found in the Appendix~\ref{app:data}.

\subsection{Multilingual Instruction Dataset}\label{sec:method_mid}
Alpaca~\cite{alpaca} contains 52K instruction and demonstration pairs generated by OpenAI's \texttt{text-davinci-003} engine using the self-instruct technique~\cite{wang-etal-2023-self-instruct}.
Some of the existing multilingual instruction datasets are primarily translated from the Alpaca datasets. For instance,~\citet{lai-etal-2023-okapi} and~\citet{li2023bactrian} expanded the Alpaca dataset to include 26 and 52 languages, respectively.
To further scale the multilingual capabilities of LLMs, we expand the Alpaca dataset to 100 languages.
Our construction process contains two steps: instruction translation and hybrid response generation.

\paragraph{Instruction Translation.}
We use Google Translate API to translate English instructions and inputs in the Alpaca dataset into 100 languages covered in the FLORES-101 dataset~\cite{goyal-etal-2022-flores}.
We chose the Google Translate API because it outperforms state-of-the-art translators such as NLLB, DeepL\footnote{\url{www.deepl.com}}, and GPT-4~\cite{achiam2023gpt} in translation performance across multiple languages~\cite{yang2023bigtrans,robinson-etal-2023-chatgpt}.
For languages that are not supported by Google Translate, we employ the NLLB model~\cite{costa2022no} for translation, as it is currently considered the state-of-the-art for multilingual translation, particularly for low-resource languages.
Similar to Bactrian-X~\cite{li2023bactrian}, we do not translate instructions that contain program-related text.

\paragraph{Hybrid Response Generation.}
Intuitively, there are two alternatives to obtain responses in various languages: the translation-based method and the generation-based method.
The translation-based approach involves directly translating Alpaca's English responses into one of the 100 target languages using either the Google Translate API or the state-of-the-art multilingual machine translation model.
Generation-based methods, like Bactrian-X~\cite{li2023bactrian}, take instructions that have been translated into the desired target language and feed them into LLMs (e.g., ChatGPT), resulting in a response expressed in the target language.
Both of these methods can produce responses in the target language, however, they also have notable limitations.
Without the context, the translation-based approach usually translates responses in a different style from the native speaker and has the potential problem of \texttt{translationese}~\cite{riley-etal-2020-translationese}.
More importantly, the understanding and generating capabilities of LLMs varies significantly across languages, and it is extremely difficult for the generation-based method to generate a high quality response in low-resource languages.
To solve the above problems, we generate responses in a hybrid mode.
We motivate our approach with the translation performance (i.e., translation task from English to other languages) of ChatGPT in the FLORES benchmark demonstrated in~\citet{lu2023chain}.
We assume that languages exhibiting poor translation performance (typically with BLEU scores below 10) by ChatGPT also possess limited generating capabilities.
Consequently, for languages with poor translation quality, we directly translate the English responses in Alpaca into the specific target language using the Google Translate API or NLLB model, while languages with good translation quality have their responses generated by ChatGPT.
ChatGPT's responses are preferable because they seem to have a more consistent style.
Table~\ref{tab:instruct_response_statistic} in the Appendix~\ref{app:data} details the translator (Google Translate API or the NLLB model) used for each language's translation instructions, as well as the method (ChatGPT, Google Translate API, or the NLLB model) employed to generate responses.

\begin{figure}[!t]
\centering
        \includegraphics[width=\linewidth]{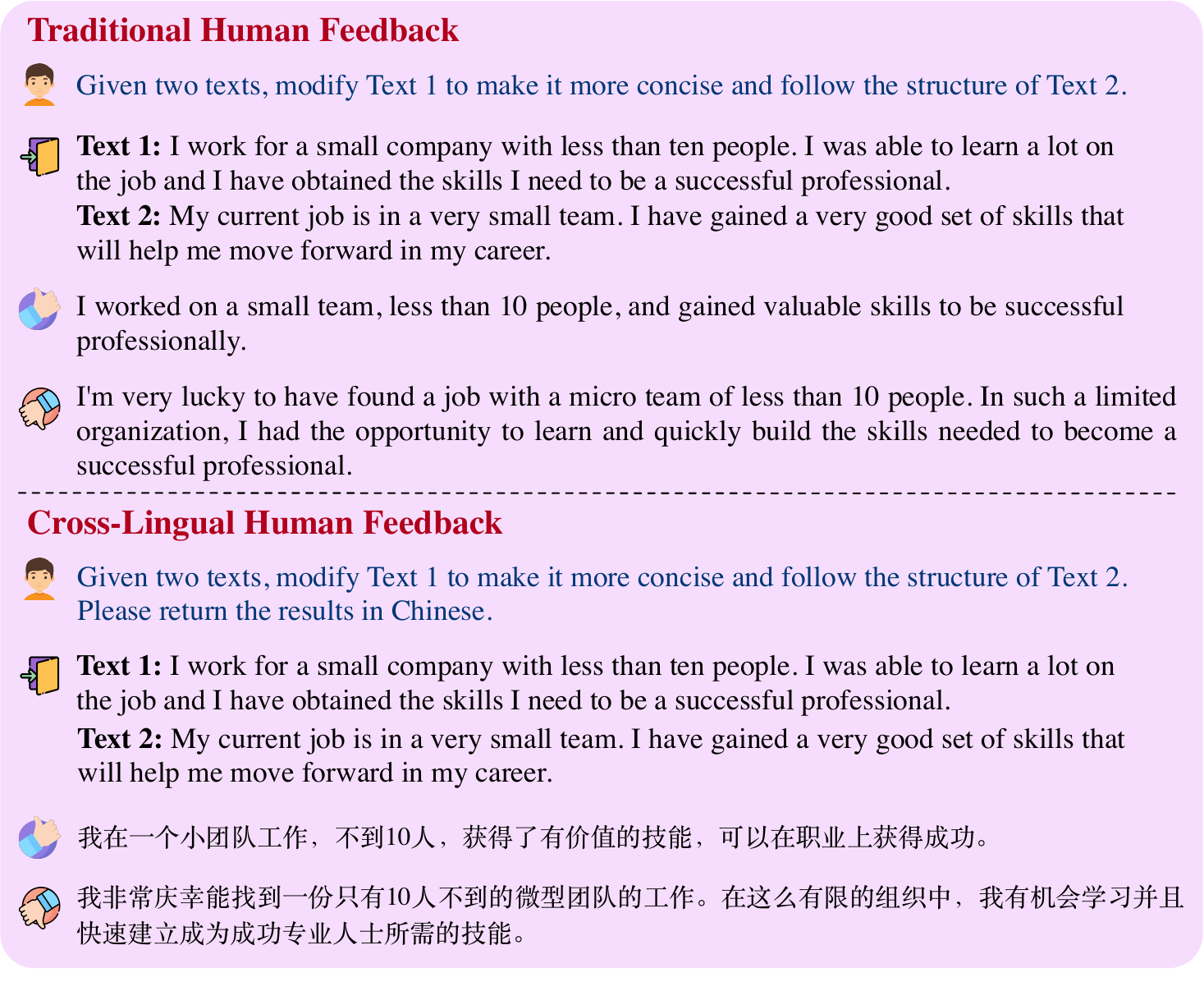}
	\caption{\label{fig:clhf}
 Cross-lingual human feedback dataset. Given instructions and inputs written in English, both the accepted and rejected outputs are written in Chinese.
 }
\end{figure}

\subsection{Cross-Lingual Feedback Dataset}\label{sec:method_clfd}
Aligning LLMs with human preferences is crucial in enhancing the truthfulness of their generated responses.
Most datasets with human feedback are monolingual, e.g., English~\cite{peng2023instruction} and Chinese~\cite{sun2023safety}.
Recently,~\citet{lai-etal-2023-okapi} extend the human feedback dataset to 26 languages using two rounds of dialogue (translation and ranking) via ChatGPT.
However, one of the biggest problems with human feedback data nowadays is that their instructions, inputs and outputs are in the same language, which limits the generative capability of LLMs.
To enhance the generating capability, we construct a cross-lingual human feedback dataset covering 30 languages. 
This dataset provides both the instruction and output in different languages.
Such a design is advantageous as it simulates a wide range of generating scenarios, thereby enhancing the generating capability of LLMs. For instance, if we have human preference data available in 30 languages, we can simulate up to $30\times29=870$ generation scenarios.
The construction process has two steps: instruction design and response generation.
We show an example generated cross-lingual human feedback  in Figure~\ref{fig:clhf}.

\paragraph{Instruction Design.}
Given a source language $\ell_{s}$, we first translate the instruction written in English into the instruction written in $\ell_{s}$ using Google Translate API.
We denote the instruction written in $\ell_{s}$ as $I^{\ell_{s}}$. 
Then, we randomly select one of the 30 languages, excluding $\ell_{s}$, as the target language $\ell_{t}$ and design an instruction written in $\ell_{s}$ to return an output written in $\ell_{t}$. 
We denote the instruction written in $\ell_{t}$ as $I^{\ell_{t}}$.
Finally, we merge $I^{\ell_{s}}$ and $I^{\ell_{t}}$ to construct a new instruction as $I_{\ell_{s}}^{\ell_{t}}$.

\paragraph{Response Generation.}
Inspired by~\citet{lai-etal-2023-okapi}, we rank the responses from ChatGPT according to their
quality for the instruction and input text.
Given the new instruction $I_{\ell_{s}}^{\ell_{t}}$, we use ChatGPT to generate the responses in target language $\ell_{t}$ and rank the responses.
The ranking process scores different responses based on three factors: correctness, coherence and naturalness.
We take the response with the highest score as the accepted response, denoted as $R_{a}$ and the response with the lowest score as the rejected response, denoted as $R_{r}$.
Note that, both the $R_{a}$ and $R_{r}$ are written in $\ell_{t}$.

\begin{table}[!ht]
\centering
\begin{tabular}{l|cc}
\toprule
           & BLEU & COMET \\
\midrule
{[}0,10)   & 2    & 0     \\
{[}10,20)  & 7    & 0     \\
{[}20,30)  & 15   & 0     \\
{[}30,40)  & 18   & 0     \\
{[}40,50)  & 26   & 3     \\
{[}50,60)  & 16   & 8     \\
(60,70{]}  & 9    & 19    \\
(70,80{]}  & 5    & 29    \\
(80,90{]}  & 2    & 32    \\
(90,100{]} & 0    & 9    \\
\bottomrule
\end{tabular}
\caption{
\label{tab:instruct_quality}
The number of languages for BLEU and COMET scores fall within each interval, obtained by back-translating from 100 languages into English.
}
\end{table}

\begin{table}[!ht]
\resizebox{\columnwidth}{!}{
\begin{tabular}{lcc||lcc}
\toprule
\multicolumn{3}{c}{\textbf{High}} & \multicolumn{3}{c}{\textbf{Low}}  \\
\cmidrule(lr){1-3}\cmidrule(lr){4-6}
         & BLEU   & CP   &           & BLEU  & CP   \\
\midrule
Arabic   & 73.16  & 0.82 & Armenian  & 47.16 & 0.64 \\
Chinese  & 80.27  & 0.91 & Gujarati  & 39.68 & 0.55 \\
French   & 77.71  & 0.85 & Kannada   & 41.72 & 0.57 \\
German   & 75.50  & 0.84 & Malayalam & 45.24 & 0.62 \\
Hindi    & 73.26  & 0.81 & Marathi   & 41.37 & 0.56 \\
\midrule
\textbf{Avg.}     & 75.98  & 0.85 & \textbf{Avg.}      & 43.03 & 0.59 \\
\bottomrule
\end{tabular}
}
\caption{
\label{tab:response_quality}
BLEU and content preservation (CP) of the response quality for 5 high-resource laguages and 5 low-resource languages.
}
\end{table}

\subsection{Instruction and Response Quality}
\label{sec:eval_dataset}
To evaluate the quality of the instructions in our constructed dataset, we randomly choose 50 instructions per language and translate them into English using the Google Translate API.
Then, we evaluate the BLEU~\cite{post-2018-call} and COMET~\cite{rei-etal-2020-comet} of the back-translated instructions against the original English instructions in Alpaca.
Table~\ref{tab:instruct_quality} shows the number of languages within each interval segment for BLEU and COMET scores.
We find that most languages have BLEU scores between $20$ and $60$ and COMET scores between $60$ and $90$, indicating that the quality of the constructed instructions is high.

To evaluate the quality of the generated responses, we randomly select $5$ high-resource (Arabic, Chinese, French, German and Hindi) and $5$ low-resource languages (Armenian, Gujarati, Kannada, Malayalam and Marathi), and evaluate 100 responses in each language.
We use two metrics for this evaluation.
First, similar to our evaluation on the instruction quality, we back-translate the responses into English and calculate the BLEU score.
Second, we assess content preservation (CP), which is a crucial metric in text style transfer~\cite{jin-etal-2022-deep}.
It measures the degree of meaning preservation between two texts by calculating the cosine similarity between the vectors of the original and generated texts.
We use this metrics by mapping the language-specific responses and the original English responses in Alpaca into the same vector space using a multilingual sentence embedding model (LaBSE;~\citealp{feng-etal-2022-language}), then we compute their cosine distance.
As shown in Table~\ref{tab:response_quality}, the BLEU scores for responses in high-resource languages ranges from $73.16$ to $80.27$ and effectively preserved meaning (i.e., CP ranges from $0.81$ to $0.91$).
In addition, the CP score is more than 0.8 for high-resource languages and about 0.6 for low-resource ones, which indicate enough quality for the purpose of this research.

\subsection{Multilingual Instruction Tuning}
\label{sec:method_training}
To enhance both the understanding and generating capabilities of LLMs, we employ a two-step training process: supervised fine-tuning and alignment with human preferences.

\paragraph{Supervised Fine-tuning (SFT).}
Starting with an LLM, e.g., LLaMA~\cite{touvron2023llama} or BLOOM~\cite{workshop2022bloom}, we perform supervised fine-tuning on the LLM using our constructed multilingual instruction dataset.
Since fine-tuning on full parameters across all layers of an LLM is computationally expensive, we apply a parameter-efficient fine-tuning technique, specifically LoRA~\cite{hu2021lora}.
LoRA incorporates trainable rank decomposition matrices into the LLM layers and only updates the newly introduced parameters while freezing all parameters of the original LLMs.
LoRA has been shown to achieve comparable performance to full parameter fine-tuning methods on LLMs~\cite{chen2023monolingual}. 

\paragraph{Aligning LLMs with human feedback.}
Common practice for aligning LLMs with human feedback is to use reinforcement learning (RLHF;~\citealp{ouyang2022training}) to employ the Proximal Policy Optimization (PPO;~\cite{schulman2017proximal}) algorithm to maximizes the reward of the model.  
However, RLHF suffers from instability of reinforcement technical and requires significant computational resources~\cite{casper2023open}.
To save computational resources, we further fine-tune the trained SFT model from the last step with the DPO algorithm~\cite{rafailov2023direct} with our constructed cross-linguistic human feedback dataset.
Note that we also use LoRA in the whole DPO training process.
\section{Experiments}
\subsection{Datasets and Tasks}
We evaluate \md on five typical benchmarks including generation, reasoning, understanding and expert-written tasks that measure the multilingual capabilities of LLMs, including both high-resource and low-resource languages.

\noindent\textbf{Understanding Task.}
We evaluate our LLM for the paraphrase identification task on PAWS-X~\cite{yang-etal-2019-paws} benchmark on 7 languages.
The benchmark provides two sentences and asks the model to determine whether they paraphrase each other or not.

\noindent\textbf{Generation Task.}
We evaluate the multilingual capability of our LLM on the FLORES-101~\cite{goyal-etal-2022-flores} and XL-Sum~\cite{hasan-etal-2021-xl} benchmarks.
FLORES-101 is a machine translation benchmark, containing parallel sentences between 101 languages. 
FLORES-101 lets us evaluate our LLM for tasks in which input and output texts are in different languages.
XL-Sum is a summarization benchmark covering 44 languages, where an LLM should summarize a long text into a short text.
XL-Sum lets us evaluate LLMs for tasks where input and output texts are in the same language.

\noindent\textbf{Reasoning Task.}
We evaluate the commonsense reasoning task using the XCOPA~\cite{ponti-etal-2020-xcopa} benchmark, which includes texts written in 11 languages.
A data sample in XCOPA consists of one premise and two choices, and requires the model to select which choice is the effect or cause of the given premise.

\noindent\textbf{Expert-written Task.}
We use the Google Translate API to translate the Self-Instruct~\cite{wang-etal-2023-self-instruct} benchmark from English to five high-resource languages (Arabic, Czech, German, Chinese, Hindi) and five low-resource languages (Armenian, Kyrgyz, Yoruba, Tamil, Mongolian). 
We call this dataset Self-Instruct*.
This benchmark contains the instruction, input and output in each instance, and requires that the LLM predicts the correct answer in the correct target language.

\subsection{Baselines}
We compare \md with the following baselines:

\noindent\textbf{Off-the-shelf LLMs.}
We evaluate LLaMA-2~\cite{touvron2023llama} and BLOOM~\cite{workshop2022bloom} as vanilla LLM baselines without  additional finetuning.

\noindent\textbf{Publicly available multilingual instruction-tuned models:}
Bactrian-X is the instruction-tuned model proposed by~\citet{li2023bactrian}. 
These models were instruction-tuned on 52 languages. 
They released models based on LLaMA and BLOOM.
We refer to them as \bxllama and \bxbloom, respectively.

\noindent\textbf{Supervised Fine-Tuning (SFT):}
We performed instruction tuning~\cite{alpaca} by utilizing our constructed multilingual instruction dataset.
We denote these models as \sftllama and \sftbloom.

\subsection{Implementation}
We use the 7B model of LLaMA (chat version;~\texttt{Llama-2-7b-chat-hf}) and BLOOM (basic version;~\texttt{bloom-7b1}) in all experiments as the base model.
We train our model using PyTorch with the HuggingFace transformers\footnote{\url{https://github.com/huggingface/transformers}} and PEFT\footnote{\url{https://github.com/huggingface/peft}} implementation.
Hyperparameters used for training our model can be found in Appendix~\ref{app:hyper_para}.
We evaluate our model in the zero-shot setting. 
Recent studies have shown that slight modifications in input prompts can lead to varied results~\cite{loya-etal-2023-exploring}.
To ensure the reproducibility, we present the prompts used for each task in the experiment in Appendix~\ref{app:prompt}. 
To mitigate over-fitting, we set the number of epochs to $1$ for both our multilingual fine-tuning and DPO training processes.

\subsection{Experimental Settings}
Similar to~\citet{huang-etal-2023-languages}, to save inference time, we randomly select $200$ test samples for each language in FLORES-101 and $250$ test samples for each language in XL-Sum.
For FLORES-101, we have two settings: translation tasks from English to other languages and translation tasks from the other languages to English, denoted as FLORES(f) and FLORES(t), respectively.  
In addition to XL-Sum and PAWS-X benchmark, we categorize the languages in each dataset into low-resource and high-resource languages based on the language classification in~\citet{costa2022no}.
We classify the languages in XL-Sum dataset into low, mid, and high categories according to~\citet{hasan-etal-2021-xl}.
The PAWS-X dataset does not include low-resource languages.

\subsection{Evaluation Metric}
For FLORES-101, we report case-sensitive detokenized BLEU with SacreBLEU\footnote{\url{https://github.com/mjpost/sacrebleu}}\cite{post-2018-call}.
For the XCOPA and PAWS-X benchmarks, we utilize the accuracy score for evaluation.
For the XL-Sum and Self-Instruct* benchmark, we report the multilingual ROUGE-1 score implemented by~\citet{lin-2004-rouge}.
\section{Results}
\label{sec:results}

\begin{table*}[!t]
\centering
\resizebox{\textwidth}{!}{
\begin{tabular}{lc|cc|cc|ccc|cc|ccc}
\toprule
\multicolumn{13}{c}{\textbf{Understanding Capabilities}} \\
\midrule
           & \textbf{PAWS-X} & \multicolumn{2}{c}{\textbf{XCOPA}} & \multicolumn{2}{c}{\textbf{Self-Instruct*}} & \multicolumn{3}{c}{\textbf{XL-Sum}} & \multicolumn{2}{c}{\textbf{FLORES(f)}} & \multicolumn{2}{c}{\textbf{FLORES(t)}} \\
           &                         & low         & high        & low             & high            & low     & mid    & high         & low           & high          & low           & high          \\
\midrule
LLaMA      & 38.10 & 47.44 & 47.22 & 7.09  & 12.57 & 4.07  & 5.44  & 2.84  & 3.07 & 4.95 & 2.96 & 6.61 \\
\bxllama   & 37.28 & 49.53 & 49.00 & 6.31  & 11.88 & 2.17  & 5.52  & 7.89  & 2.69 & 2.38 & 3.15 & 5.31 \\
\sftllama  & 42.32 & 50.19 & 49.86 & 7.32  & 12.72 & 4.70  & 7.34  & 7.55  & 3.13 & 3.93 & 3.16 & 6.92 \\
\md        & \textbf{46.95} & \textbf{51.53} & \textbf{51.96} & \textbf{12.94} & \textbf{15.35} & \textbf{8.83}  & \textbf{13.90} & \textbf{17.29} & \textbf{3.27} & \textbf{8.09} & \textbf{4.04} & \textbf{14.18}  \\
\midrule
BLOOM      & 36.47 & 44.27 & 49.14 & 7.56  & 8.67  & 9.03  & 14.06 & 16.80 & 2.54 & 2.04 & 2.05 & 2.56 \\
\bxbloom   & 36.42 & 46.28 & 50.35 & 4.81  & 8.11  & 4.89  & 8.47  & 11.71 & 2.14 & 1.74 & 2.41 & 1.57 \\
\sftbloom  & 36.67 & 49.42 & 52.31 & 6.31  & 11.88 & 5.62  & 10.12 & 14.33 & \textbf{3.12} & 3.79 & 2.62 & 2.52 \\
\md        & \textbf{39.83} & \textbf{52.50} & \textbf{55.59} & \textbf{7.94}  & \textbf{13.35} & \textbf{12.87} & \textbf{15.23} & \textbf{18.38} & 3.02 & \textbf{4.71} & \textbf{3.94} & \textbf{6.54}  \\
\midrule
\multicolumn{13}{c}{\textbf{Generating Capabilities}} \\
\midrule
           & \textbf{PAWS-X} & \multicolumn{2}{c}{\textbf{XCOPA}} & \multicolumn{2}{c}{\textbf{Self-Instruct*}} & \multicolumn{3}{c}{\textbf{XL-Sum}} & \multicolumn{2}{c}{\textbf{FLORES(f)}} & \multicolumn{2}{c}{\textbf{FLORES(t)}} \\
           &                         & low         & high        & low             & high            & low    & mid    & high         & low           & high          & low           & high          \\
\midrule
LLaMA      & 50.22 & 49.33 & 51.52 & 5.38  & 8.81  & 6.26  & 5.80  & 8.08  & 1.35 & 3.90 & 2.11 & 4.95 \\
\bxllama   & 48.41 & 48.00 & 49.85 & 7.01  & 9.80  & 1.11  & 2.74  & 1.70  & 1.56 & 5.33 & 1.37 & 1.61 \\
\sftllama  & 50.36 & 48.93 & 50.05 & 7.10  & 12.15 & 4.51  & 6.06  & 9.21  & 2.42 & 4.56 & 2.71 & 7.29  \\
\md        & \textbf{61.94} & \textbf{49.71} & \textbf{54.68} & \textbf{9.16}  & \textbf{14.71} & \textbf{9.99}  & \textbf{13.57} & \textbf{16.61} & \textbf{2.89} & \textbf{9.07} & \textbf{5.64} & \textbf{16.98} \\
\midrule
BLOOM      & 47.39 & 49.85 & 49.47 & 4.07  & 7.01  & 6.08  & 7.77  & 8.91  & 0.78 & 1.20 & 0.99 & 1.49 \\
\bxbloom   & 47.26 & 47.72 & 49.98 & 5.88  & 8.21  & 1.98  & 3.59  & 4.58  & 0.47 & 0.82 & 1.95 & 2.33   \\
\sftbloom  & 48.50 & 49.13 & 49.28 & 7.78  & 11.51 & 3.89  & 8.87  & 10.89 & 2.59 & 3.12 & 2.05 & 2.56   \\
\md        & \textbf{50.53} & \textbf{52.36} & \textbf{52.26} & \textbf{10.17} & \textbf{13.62} & \textbf{8.77}  & \textbf{11.74} & \textbf{12.36} & \textbf{3.97} & \textbf{5.79} &\textbf{ 4.22} & \textbf{7.68}  \\
\bottomrule
\end{tabular}}
\caption{\label{tab:main_res}
\textbf{Understanding and generating capability of LLMs.} We evaluate on five benchmarks covers both on high-resource and low-resource languages. We utilize accuracy score to evaluate on PAWS-X and XCOPA. We evaluate Self-Instruct* and XL-Sum using ROUGE-1 score and FLORES using BLEU.
}
\end{table*}

Our goal is to simultaneously evaluate the understanding capability and the generating capability of the LLMs.
To evaluate the understanding capability of LLMs, we use Google Translate API to translate the instructions for each task into different languages. 
To evaluate the generating capability of an LLM, we use English as the instruction language during the inference phase.
We choose English because LLMs demonstrate effective comprehension of English instructions, thus avoiding any potential impact on the generating capability of LLMs due to misunderstanding of instructions.
Table~\ref{tab:main_res} presents the average score for all languages in each benchmark.
For more comprehensive results of each benchmark per language, please refer to Appendix~\ref{app:full_res}.

The off-the-shelf LLMs demonstrate limited multilingual capabilities, in particular in generation tasks, where performance is exceedingly poor.
After fine-tuning the LLMs using the multilingual instruction dataset, there is a marginal (but not significant) improvement in the multilingual capability of the LLMs across most benchmarks, especially when describing instructions in non-English languages.
For example, the understanding capability of the BX-based models (\bxllama and \bxbloom) drops significantly on the Self-Instruct*, XL-Sum, and FLORES benchmarks. 
This phenomenon is due to the fact that the BX-based models are fine-tuned in 52 languages and do not completely support all the languages in the benchmarks. 

\textbf{Low-Resource vs High-Resource.}
We observe that the multilingual capabilities of the examined LLMs are significantly superior in high-resource languages to those in low-resource languages.
It is a common problem for multilingual models that the scarcity of training corpora in low-resource languages makes it challenging to train a robust decoder, leading to the generation of incorrect outputs~\cite{lai-etal-2023-mitigating}.

\textbf{Understanding Capability vs Generating Capability.}
Our findings indicate that instructions written in English outperform instructions written in non-English languages, aligning with our initial expectations.
This phenomenon can be attributed to the accumulation of biases in LLMs' understanding of instructions across different languages.
Such biases can introduce errors during the generation of results, resulting in incorrect outputs or outputs in the wrong language (i.e., off-target problem;~\citealp{zhang-etal-2020-improving}).

\textbf{\md.}
Compared with the baseline models, our models demonstrate consistent improvements for both multilingual capabilities across all examined tasks.
In comparison to the SFT-based models, \md increases the step of alignment with human preferences, resulting in additional improvements. 
This observation highlights the significance of aligning with human preferences after supervised fine-tuning.
Furthermore, our model significantly outperforms other models on generative task datasets such as XL-Sum and FLORES, showing the effectiveness of our dataset and human feedback finetuning.
\section{Analysis}
In Section \ref{sec:results} we show the effectiveness of our approach compared with  previous work. In this section, we study the performance of our multilingual model in  detail.

\begin{table}[!t]
\begin{tabular}{lrrrr}
\toprule
              & \multicolumn{2}{c}{\textbf{Low}} & \multicolumn{2}{c}{\textbf{High}} \\
\cmidrule(lr){2-3} \cmidrule(lr){4-5}
              & mono       & cross               & mono        & cross      \\
\midrule
PAWS-X        & -          & -                   & 58.43       & \textbf{61.94}      \\
XCOPA         & 47.26      & \textbf{49.71}      & 52.15       & \textbf{54.68}      \\
Self-Instruct* & 3.25       & \textbf{9.16}       & 12.14       & \textbf{14.71}      \\
XL-Sum        & 3.38       & \textbf{9.99}       & 12.52       & \textbf{16.61}      \\
FLORES(f)     & 0.85       & \textbf{2.89}       & 4.57        & \textbf{9.07}       \\
FLORES(t)     & 1.55       & \textbf{5.64}       & 8.45        & \textbf{16.98}      \\
\bottomrule
\end{tabular}
\caption{\label{tab:ablation}
An ablation study of \md using mono-lingual and cross-lingual human feedback data on low- and high-resource languages.
}
\end{table}

\subsection{Different Human Feedback Datasets}
To investigate the importance of cross-lingual properties in aligning LLMs with human preferences, we conduct an ablation study on the same five benchmarks as shown in Table~\ref{tab:main_res}.
In particular, we employ the DPO algorithm~\cite{rafailov2023direct} to finetune our model,~\md, on two distinct datasets.
The first dataset is our constructed cross-lingual human feedback dataset, where instructions and outputs are in different languages.
The second dataset is a traditional monolingual human feedback dataset~\cite{lai-etal-2023-okapi}, where both instructions and outputs are in the same language.
Table~\ref{tab:ablation} show the results categorized by low- and high-resource languages.
We observe that 
(1) Aligning \md using our cross-lingual human feedback dataset yields superior results compared with using monolingual human feedback. 
This improvement is evident for datasets with generation tasks such as XL-SUM and FLORES, showing that our novel cross-lingual human feedback dataset effectively simulates the multilingual generation task, reducing the possibilities of generating incorrect outputs.
(2) Finetuning \md on our cross-lingual human feedback dataset is more effective for low-resource languages than high-resource ones.  
This is due to the fact that high-resource languages already exhibit strong understanding and generation capabilities in the vanilla LLMs (as shown in Table~\ref{tab:main_res}), which mitigates the impact of further finetuning \md on the cross-lingual preference data. 
(3) It is worth noting that despite aligning \md with cross-lingual human preferences, its performance in low-resource languages is still not as good as that in high-resource languages. 
Although our constructed cross-lingual feedback dataset enhances multilingual performance, the inclusion of additional languages (our dataset currently includes 30 languages) might be necessary to support all low-resource languages in the evaluation benchmarks.

\begin{table}[t]
\resizebox{\columnwidth}{!}{
\begin{tabular}{lrrrr}
\toprule
              & \multicolumn{2}{c}{\textbf{Low}} & \multicolumn{2}{c}{\textbf{High}} \\
\cmidrule(lr){2-3} \cmidrule(lr){4-5}
              & para      & instruct    & para      & instruct     \\
\midrule
PAWS-X        & -         & -           & 40.17     & \textbf{50.36}        \\
XCOPA         & 37.14     & \textbf{48.93}       & 42.13     & \textbf{50.05}        \\
Self-Instruct* & 2.63      & \textbf{7.10}        & 5.48      & \textbf{12.15}        \\
XL-Sum        & 1.10      & \textbf{4.51}        & 5.12      & \textbf{9.21}         \\
FLORES(f)     & \textbf{5.06}      & 2.42        & \textbf{13.27}     & 4.56         \\
FLORES(t)     & \textbf{12.36}     & 2.71        & \textbf{18.27}     & 7.29         \\
\bottomrule
\end{tabular}
}
\caption{\label{tab:data_compare}
     Multilingual Tuning on multilingual parallel corpora and multilingual instruction dataset in high-resource and low-resource languages.
}
\end{table}

\begin{table}[t]
\centering
\begin{tabular}{lrrrr}
\toprule
         & \multicolumn{2}{c}{\textbf{FLORES(f)}} & \multicolumn{2}{c}{\textbf{FLORES(t)}} \\
\cmidrule(lr){2-3} \cmidrule(lr){4-5}
         & Low           & High          & Low           & High          \\
\midrule
LLaMA    & 23.26         & 16.76         & 14.15         & 10.16         \\
\bxllama  & 14.13         & 8.32          & 12.17         & 8.24          \\
\sftllama & 10.26         & 6.34          & 8.72          & 6.23          \\
\md       & \textbf{8.82}          & \textbf{3.47}          & \textbf{6.95}          & \textbf{1.46}   \\
\bottomrule
\end{tabular}
\caption{\label{tab:off_target}
         OTR scores (lower is 
         better) of examined multilingual LLMs on the 
         FLORES
         benchmark.
 }
\end{table}

\begin{table}[!t]
\centering
\resizebox{\columnwidth}{!}{
\begin{tabular}{lrrrr}
\toprule
              & \textbf{LLaMA} & \textbf{\bxllama} & \textbf{\sftllama} & \textbf{\md}    \\
\midrule
PAWS-X        & 60.56 & 58.77   & 60.63    & \textbf{66.43} \\
XCOPA         & 93.33 & 98.52   & \textbf{99.31}    & 89.63 \\
Self-Instruct* & 57.85 & 68.68   & 62.63    & \textbf{73.92} \\
XL-Sum        & 47.09 & 8.90    & 50.35    & \textbf{67.21} \\
FLORES(f)     & 34.33 & 34.00   & 25.84    & \textbf{34.68} \\
FLORES(t)     & 49.84 & \textbf{58.28}   & 35.53    & 48.28 \\
\bottomrule
\end{tabular}}
\caption{\label{fig:decom}
        \textbf{Language Democratization:} Mitigating the gap between the average performance and the best performances of each task in different languages.
}
\end{table}

\subsection{Different Datasets for Multilingual Tuning}
\label{sec:diff_data}
In Section~\ref{sec:intro}, we introduced two types of datasets to enhance the multilingual capabilities of LLMs: multilingual parallel corpus and multilingual instruction dataset.
We study how these two types of data impact the multilingual capabilities of LLMs.
To do so, we conduct comprehensive comparison experiments on these two types of dataset.
For the multilingual parallel corpus, we use the NLLB dataset~\cite{costa2022no} collected by allenai\footnote{\url{https://huggingface.co/datasets/allenai/nllb}}.
We use the same set of 100 languages as \md for the sake of a fair comparison. 

The experimental results presented in Table~\ref{tab:data_compare} clearly show that utilizing the multilingual parallel corpus leads to a significant improvement in the machine translation task. 
However, there is a notable decrease in performance observed in non machine translation tasks. 
This phenomenon is referred to as catastrophic forgetting~\cite{mccloskey1989catastrophic}, where the model after fine-tuning achieves better performance on the new task at the expense of the model's performance on other tasks.
On the other hand, the model fine-tuned with multilingual instruction data demonstrates improvement across all tasks, indicating that attention should be given to the models' performance on a broad range of tasks during fine-tuning, rather than focusing solely on performance gains in a single task.
It is obvious that the multilingual instruction dataset is a better choice compare to the multilingual parallel dataset when finetuning LLMs.

\subsection{Off-Target Analysis}
Off-target \cite{zhang-etal-2020-improving} refers to the generation of output in an incorrect language, which is a common issue in multilingual models. 
Following the approach of \citet{lai2023extending}, we calculate the Off-Target Ratio (OTR), which represents the proportion of output sentences generated by a multilingual model that are in the wrong language.
This metric helps to assess the accuracy of language generation in multilingual models.
Table~\ref{tab:off_target} shows the results on the XL-Sum and FLORES benchmark.
We observe that the off-target problem is more prominent in low-resource languages. 
Although our model has made some progress in addressing this issue, there is still significant room for improvement. Overcoming the off-target problem in low-resource languages continues to be a challenge that necessitates further research and development efforts.

\subsection{Language Democratization}
Language democratization, as proposed by~\citet{huang-etal-2023-languages}, is a metric used to evaluate the level of task democratization across different languages of a multilingual model.
This metric is obtained by calculating the average percentage of different languages relative to the best performing language among all languages.
It provides insights into the fairness and equality of performance across different languages in a multilingual model.
According to Table~\ref{fig:decom}, we observe that \md demonstrates a higher degree of linguistic democratization its peers across four out of six examined benchmarks.  
This implies that our model exhibits a smaller performance gap between different languages, indicating a more equal and fair distribution of performance across languages.
This is a positive outcome, suggesting that our model is successful in reducing disparities and achieving more balanced performance across various languages.

\section{Conclusions}
To enhance the multilingual capability of LLMs in two dimensions (understanding and generating), we present \md, two LLMs that support 100 languages.
We train \md on a novel multilingual instruction dataset containing 100 languages.
To improve the generating capability, we construct a cross-lingual human feedback dataset to further align the LLMs with human feedback and to enable the LLMs to generate output in multiple languages.
Experiments on five benchmarks demonstrate the effectiveness of our datasets and also the multilingual capability of our models both on high-resource languages and low-resource languages.
\section{Limitations}
This work has the following limitations:
(i) To make our computations environment friendly, our experiments have so far been limited to 7B size on LLaMA and BLOOM. 
However, our dataset has the potential to be deployed for larger LLMs (e.g., 13B and 70B models). 
We hope that the community will contribute to realizing this potential using our dataset.
(ii) Our constructed human feedback dataset currently covers 30 languages. 
Given that ChatGPT also faces difficulties in obtaining high-quality feedback across a majority of languages, determining how to extend the cross-lingual human feedback dataset would be promising future work.
(iii) As shown in Appendix~\ref{app:multi_intro} and~\ref{app:slms}, a noticeable discrepancy persists between \md, small language models (SLMs) and the current state-of-the-art LLMs, such as ChatGPT.
This disparity is primarily due to models like ChatGPT leveraging larger quantities of data and model sizes. 
Therefore, bridging this gap will be a critical objective in future research.
(iv) We do not conduct an analysis on toxicity~\cite{deshpande-etal-2023-toxicity}, domain~\cite{lai-etal-2022-m4,lai-etal-2022-improving-domain}, bias and fairness~\cite{gallegos2023bias} aspects of \md, which should be discussed more in future work.

\section*{Acknowledgement}
This publication was partially supported by LMUexcellent, funded by the Federal Ministry of Education and Research (BMBF) and the Free State of Bavaria under the Excellence Strategy of the Federal Government and the Länder; and by the German Research Foundation (DFG; grant FR 2829/4-1).

\bibliography{anthology,custom}

\begin{thebibliography}{48}
\expandafter\ifx\csname natexlab\endcsname\relax\def\natexlab#1{#1}\fi

\bibitem[{Achiam et~al.(2023)Achiam, Adler, Agarwal, Ahmad, Akkaya, Aleman, Almeida, Altenschmidt, Altman, Anadkat et~al.}]{achiam2023gpt}
Josh Achiam, Steven Adler, Sandhini Agarwal, Lama Ahmad, Ilge Akkaya, Florencia~Leoni Aleman, Diogo Almeida, Janko Altenschmidt, Sam Altman, Shyamal Anadkat, et~al. 2023.
\newblock Gpt-4 technical report.
\newblock \emph{arXiv preprint arXiv:2303.08774}.

\bibitem[{AI@Meta(2024)}]{llama3modelcard}
AI@Meta. 2024.
\newblock \href {https://github.com/meta-llama/llama3/blob/main/MODEL_CARD.md} {Llama 3 model card}.

\bibitem[{Artetxe et~al.(2020)Artetxe, Ruder, and Yogatama}]{artetxe-etal-2020-cross}
Mikel Artetxe, Sebastian Ruder, and Dani Yogatama. 2020.
\newblock \href {https://doi.org/10.18653/v1/2020.acl-main.421} {On the cross-lingual transferability of monolingual representations}.
\newblock In \emph{Proceedings of the 58th Annual Meeting of the Association for Computational Linguistics}, pages 4623--4637, Online. Association for Computational Linguistics.

\bibitem[{BigScience et~al.(2022)BigScience, Scao, Fan, Akiki, Pavlick, Ili{\'c}, Hesslow, Castagn{\'e}, Luccioni, Yvon et~al.}]{workshop2022bloom}
Workshop BigScience, Teven~Le Scao, Angela Fan, Christopher Akiki, Ellie Pavlick, Suzana Ili{\'c}, Daniel Hesslow, Roman Castagn{\'e}, Alexandra~Sasha Luccioni, Fran{\c{c}}ois Yvon, et~al. 2022.
\newblock Bloom: A 176b-parameter open-access multilingual language model.
\newblock \emph{arXiv preprint arXiv:2211.05100}.

\bibitem[{Casper et~al.(2023)Casper, Davies, Shi, Gilbert, Scheurer, Rando, Freedman, Korbak, Lindner, Freire et~al.}]{casper2023open}
Stephen Casper, Xander Davies, Claudia Shi, Thomas~Krendl Gilbert, J{\'e}r{\'e}my Scheurer, Javier Rando, Rachel Freedman, Tomasz Korbak, David Lindner, Pedro Freire, et~al. 2023.
\newblock Open problems and fundamental limitations of reinforcement learning from human feedback.
\newblock \emph{arXiv preprint arXiv:2307.15217}.

\bibitem[{Chen et~al.(2023)Chen, Ji, Bogoychev, Haddow, and Heafield}]{chen2023monolingual}
Pinzhen Chen, Shaoxiong Ji, Nikolay Bogoychev, Barry Haddow, and Kenneth Heafield. 2023.
\newblock Monolingual or multilingual instruction tuning: Which makes a better alpaca.
\newblock \emph{arXiv preprint arXiv:2309.08958}.

\bibitem[{Costa-juss{\`a} et~al.(2022)Costa-juss{\`a}, Cross, {\c{C}}elebi, Elbayad, Heafield, Heffernan, Kalbassi, Lam, Licht, Maillard et~al.}]{costa2022no}
Marta~R Costa-juss{\`a}, James Cross, Onur {\c{C}}elebi, Maha Elbayad, Kenneth Heafield, Kevin Heffernan, Elahe Kalbassi, Janice Lam, Daniel Licht, Jean Maillard, et~al. 2022.
\newblock No language left behind: Scaling human-centered machine translation.
\newblock \emph{arXiv preprint arXiv:2207.04672}.

\bibitem[{Deshpande et~al.(2023)Deshpande, Murahari, Rajpurohit, Kalyan, and Narasimhan}]{deshpande-etal-2023-toxicity}
Ameet Deshpande, Vishvak Murahari, Tanmay Rajpurohit, Ashwin Kalyan, and Karthik Narasimhan. 2023.
\newblock \href {https://doi.org/10.18653/v1/2023.findings-emnlp.88} {Toxicity in chatgpt: Analyzing persona-assigned language models}.
\newblock In \emph{Findings of the Association for Computational Linguistics: EMNLP 2023}, pages 1236--1270, Singapore. Association for Computational Linguistics.

\bibitem[{Etxaniz et~al.(2023)Etxaniz, Azkune, Soroa, de~Lacalle, and Artetxe}]{etxaniz2023multilingual}
Julen Etxaniz, Gorka Azkune, Aitor Soroa, Oier~Lopez de~Lacalle, and Mikel Artetxe. 2023.
\newblock Do multilingual language models think better in english?
\newblock \emph{arXiv preprint arXiv:2308.01223}.

\bibitem[{Feng et~al.(2022)Feng, Yang, Cer, Arivazhagan, and Wang}]{feng-etal-2022-language}
Fangxiaoyu Feng, Yinfei Yang, Daniel Cer, Naveen Arivazhagan, and Wei Wang. 2022.
\newblock \href {https://doi.org/10.18653/v1/2022.acl-long.62} {Language-agnostic {BERT} sentence embedding}.
\newblock In \emph{Proceedings of the 60th Annual Meeting of the Association for Computational Linguistics (Volume 1: Long Papers)}, pages 878--891, Dublin, Ireland. Association for Computational Linguistics.

\bibitem[{Gallegos et~al.(2023)Gallegos, Rossi, Barrow, Tanjim, Kim, Dernoncourt, Yu, Zhang, and Ahmed}]{gallegos2023bias}
Isabel~O Gallegos, Ryan~A Rossi, Joe Barrow, Md~Mehrab Tanjim, Sungchul Kim, Franck Dernoncourt, Tong Yu, Ruiyi Zhang, and Nesreen~K Ahmed. 2023.
\newblock Bias and fairness in large language models: A survey.
\newblock \emph{arXiv preprint arXiv:2309.00770}.

\bibitem[{Goyal et~al.(2022)Goyal, Gao, Chaudhary, Chen, Wenzek, Ju, Krishnan, Ranzato, Guzm{\'a}n, and Fan}]{goyal-etal-2022-flores}
Naman Goyal, Cynthia Gao, Vishrav Chaudhary, Peng-Jen Chen, Guillaume Wenzek, Da~Ju, Sanjana Krishnan, Marc{'}Aurelio Ranzato, Francisco Guzm{\'a}n, and Angela Fan. 2022.
\newblock \href {https://doi.org/10.1162/tacl_a_00474} {The {F}lores-101 evaluation benchmark for low-resource and multilingual machine translation}.
\newblock \emph{Transactions of the Association for Computational Linguistics}, 10:522--538.

\bibitem[{Groeneveld et~al.(2024)Groeneveld, Beltagy, Walsh, Bhagia, Kinney, Tafjord, Jha, Ivison, Magnusson, Wang et~al.}]{groeneveld2024olmo}
Dirk Groeneveld, Iz~Beltagy, Pete Walsh, Akshita Bhagia, Rodney Kinney, Oyvind Tafjord, Ananya~Harsh Jha, Hamish Ivison, Ian Magnusson, Yizhong Wang, et~al. 2024.
\newblock Olmo: Accelerating the science of language models.
\newblock \emph{arXiv preprint arXiv:2402.00838}.

\bibitem[{Hasan et~al.(2021)Hasan, Bhattacharjee, Islam, Mubasshir, Li, Kang, Rahman, and Shahriyar}]{hasan-etal-2021-xl}
Tahmid Hasan, Abhik Bhattacharjee, Md.~Saiful Islam, Kazi Mubasshir, Yuan-Fang Li, Yong-Bin Kang, M.~Sohel Rahman, and Rifat Shahriyar. 2021.
\newblock \href {https://doi.org/10.18653/v1/2021.findings-acl.413} {{XL}-sum: Large-scale multilingual abstractive summarization for 44 languages}.
\newblock In \emph{Findings of the Association for Computational Linguistics: ACL-IJCNLP 2021}, pages 4693--4703, Online. Association for Computational Linguistics.

\bibitem[{Hu et~al.(2021)Hu, Wallis, Allen-Zhu, Li, Wang, Wang, Chen et~al.}]{hu2021lora}
Edward~J Hu, Phillip Wallis, Zeyuan Allen-Zhu, Yuanzhi Li, Shean Wang, Lu~Wang, Weizhu Chen, et~al. 2021.
\newblock Lora: Low-rank adaptation of large language models.
\newblock In \emph{International Conference on Learning Representations}.

\bibitem[{Huang et~al.(2023)Huang, Tang, Zhang, Zhao, Song, Xia, and Wei}]{huang-etal-2023-languages}
Haoyang Huang, Tianyi Tang, Dongdong Zhang, Xin Zhao, Ting Song, Yan Xia, and Furu Wei. 2023.
\newblock \href {https://doi.org/10.18653/v1/2023.findings-emnlp.826} {Not all languages are created equal in {LLM}s: Improving multilingual capability by cross-lingual-thought prompting}.
\newblock In \emph{Findings of the Association for Computational Linguistics: EMNLP 2023}, pages 12365--12394, Singapore. Association for Computational Linguistics.

\bibitem[{Jin et~al.(2022)Jin, Jin, Hu, Vechtomova, and Mihalcea}]{jin-etal-2022-deep}
Di~Jin, Zhijing Jin, Zhiting Hu, Olga Vechtomova, and Rada Mihalcea. 2022.
\newblock \href {https://doi.org/10.1162/coli_a_00426} {Deep learning for text style transfer: A survey}.
\newblock \emph{Computational Linguistics}, 48(1):155--205.

\bibitem[{Lai et~al.(2023{\natexlab{a}})Lai, Nguyen, Ngo, Nguyen, Dernoncourt, Rossi, and Nguyen}]{lai-etal-2023-okapi}
Viet Lai, Chien Nguyen, Nghia Ngo, Thuat Nguyen, Franck Dernoncourt, Ryan Rossi, and Thien Nguyen. 2023{\natexlab{a}}.
\newblock \href {https://doi.org/10.18653/v1/2023.emnlp-demo.28} {Okapi: Instruction-tuned large language models in multiple languages with reinforcement learning from human feedback}.
\newblock In \emph{Proceedings of the 2023 Conference on Empirical Methods in Natural Language Processing: System Demonstrations}, pages 318--327, Singapore. Association for Computational Linguistics.

\bibitem[{Lai et~al.(2022{\natexlab{a}})Lai, Chronopoulou, and Fraser}]{lai-etal-2022-m4}
Wen Lai, Alexandra Chronopoulou, and Alexander Fraser. 2022{\natexlab{a}}.
\newblock \href {https://doi.org/10.18653/v1/2022.findings-emnlp.315} {m$^4$ adapter: Multilingual multi-domain adaptation for machine translation with a meta-adapter}.
\newblock In \emph{Findings of the Association for Computational Linguistics: EMNLP 2022}, pages 4282--4296, Abu Dhabi, United Arab Emirates. Association for Computational Linguistics.

\bibitem[{Lai et~al.(2023{\natexlab{b}})Lai, Chronopoulou, and Fraser}]{lai-etal-2023-mitigating}
Wen Lai, Alexandra Chronopoulou, and Alexander Fraser. 2023{\natexlab{b}}.
\newblock \href {https://doi.org/10.18653/v1/2023.findings-emnlp.953} {Mitigating data imbalance and representation degeneration in multilingual machine translation}.
\newblock In \emph{Findings of the Association for Computational Linguistics: EMNLP 2023}, pages 14279--14294, Singapore. Association for Computational Linguistics.

\bibitem[{Lai et~al.(2023{\natexlab{c}})Lai, Hangya, and Fraser}]{lai2023extending}
Wen Lai, Viktor Hangya, and Alexander Fraser. 2023{\natexlab{c}}.
\newblock Extending multilingual machine translation through imitation learning.
\newblock \emph{arXiv preprint arXiv:2311.08538}.

\bibitem[{Lai et~al.(2022{\natexlab{b}})Lai, Libovick{\'y}, and Fraser}]{lai-etal-2022-improving-domain}
Wen Lai, Jind{\v{r}}ich Libovick{\'y}, and Alexander Fraser. 2022{\natexlab{b}}.
\newblock \href {https://aclanthology.org/2022.coling-1.461} {Improving both domain robustness and domain adaptability in machine translation}.
\newblock In \emph{Proceedings of the 29th International Conference on Computational Linguistics}, pages 5191--5204, Gyeongju, Republic of Korea. International Committee on Computational Linguistics.

\bibitem[{Li et~al.(2023)Li, Koto, Wu, Aji, and Baldwin}]{li2023bactrian}
Haonan Li, Fajri Koto, Minghao Wu, Alham~Fikri Aji, and Timothy Baldwin. 2023.
\newblock Bactrian-x: A multilingual replicable instruction-following model with low-rank adaptation.
\newblock \emph{arXiv preprint arXiv:2305.15011}.

\bibitem[{Lin(2004)}]{lin-2004-rouge}
Chin-Yew Lin. 2004.
\newblock \href {https://aclanthology.org/W04-1013} {{ROUGE}: A package for automatic evaluation of summaries}.
\newblock In \emph{Text Summarization Branches Out}, pages 74--81, Barcelona, Spain. Association for Computational Linguistics.

\bibitem[{Loya et~al.(2023)Loya, Sinha, and Futrell}]{loya-etal-2023-exploring}
Manikanta Loya, Divya Sinha, and Richard Futrell. 2023.
\newblock \href {https://doi.org/10.18653/v1/2023.findings-emnlp.241} {Exploring the sensitivity of {LLM}s{'} decision-making capabilities: Insights from prompt variations and hyperparameters}.
\newblock In \emph{Findings of the Association for Computational Linguistics: EMNLP 2023}, pages 3711--3716, Singapore. Association for Computational Linguistics.

\bibitem[{Lu et~al.(2023)Lu, Huang, Zhang, Yang, Lam, and Wei}]{lu2023chain}
Hongyuan Lu, Haoyang Huang, Dongdong Zhang, Haoran Yang, Wai Lam, and Furu Wei. 2023.
\newblock Chain-of-dictionary prompting elicits translation in large language models.
\newblock \emph{arXiv preprint arXiv:2305.06575}.

\bibitem[{Luo et~al.(2023)Luo, Kong, Xu, Cao, Hao, Qu, Chen, Zhu, Zhao, Zhang et~al.}]{luo2023yayi}
Yin Luo, Qingchao Kong, Nan Xu, Jia Cao, Bao Hao, Baoyu Qu, Bo~Chen, Chao Zhu, Chenyang Zhao, Donglei Zhang, et~al. 2023.
\newblock Yayi 2: Multilingual open-source large language models.
\newblock \emph{arXiv preprint arXiv:2312.14862}.

\bibitem[{McCloskey and Cohen(1989)}]{mccloskey1989catastrophic}
Michael McCloskey and Neal~J Cohen. 1989.
\newblock Catastrophic interference in connectionist networks: The sequential learning problem.
\newblock In \emph{Psychology of learning and motivation}, volume~24, pages 109--165. Elsevier.

\bibitem[{Muennighoff et~al.(2023)Muennighoff, Wang, Sutawika, Roberts, Biderman, Le~Scao, Bari, Shen, Yong, Schoelkopf, Tang, Radev, Aji, Almubarak, Albanie, Alyafeai, Webson, Raff, and Raffel}]{muennighoff-etal-2023-crosslingual}
Niklas Muennighoff, Thomas Wang, Lintang Sutawika, Adam Roberts, Stella Biderman, Teven Le~Scao, M~Saiful Bari, Sheng Shen, Zheng~Xin Yong, Hailey Schoelkopf, Xiangru Tang, Dragomir Radev, Alham~Fikri Aji, Khalid Almubarak, Samuel Albanie, Zaid Alyafeai, Albert Webson, Edward Raff, and Colin Raffel. 2023.
\newblock \href {https://doi.org/10.18653/v1/2023.acl-long.891} {Crosslingual generalization through multitask finetuning}.
\newblock In \emph{Proceedings of the 61st Annual Meeting of the Association for Computational Linguistics (Volume 1: Long Papers)}, pages 15991--16111, Toronto, Canada. Association for Computational Linguistics.

\bibitem[{Ouyang et~al.(2022)Ouyang, Wu, Jiang, Almeida, Wainwright, Mishkin, Zhang, Agarwal, Slama, Ray et~al.}]{ouyang2022training}
Long Ouyang, Jeffrey Wu, Xu~Jiang, Diogo Almeida, Carroll Wainwright, Pamela Mishkin, Chong Zhang, Sandhini Agarwal, Katarina Slama, Alex Ray, et~al. 2022.
\newblock Training language models to follow instructions with human feedback.
\newblock \emph{Advances in Neural Information Processing Systems}, 35:27730--27744.

\bibitem[{Peng et~al.(2023)Peng, Li, He, Galley, and Gao}]{peng2023instruction}
Baolin Peng, Chunyuan Li, Pengcheng He, Michel Galley, and Jianfeng Gao. 2023.
\newblock Instruction tuning with gpt-4.
\newblock \emph{arXiv preprint arXiv:2304.03277}.

\bibitem[{Ponti et~al.(2020)Ponti, Glava{\v{s}}, Majewska, Liu, Vuli{\'c}, and Korhonen}]{ponti-etal-2020-xcopa}
Edoardo~Maria Ponti, Goran Glava{\v{s}}, Olga Majewska, Qianchu Liu, Ivan Vuli{\'c}, and Anna Korhonen. 2020.
\newblock \href {https://doi.org/10.18653/v1/2020.emnlp-main.185} {{XCOPA}: A multilingual dataset for causal commonsense reasoning}.
\newblock In \emph{Proceedings of the 2020 Conference on Empirical Methods in Natural Language Processing (EMNLP)}, pages 2362--2376, Online. Association for Computational Linguistics.

\bibitem[{Post(2018)}]{post-2018-call}
Matt Post. 2018.
\newblock \href {https://doi.org/10.18653/v1/W18-6319} {A call for clarity in reporting {BLEU} scores}.
\newblock In \emph{Proceedings of the Third Conference on Machine Translation: Research Papers}, pages 186--191, Brussels, Belgium. Association for Computational Linguistics.

\bibitem[{Rafailov et~al.(2023)Rafailov, Sharma, Mitchell, Ermon, Manning, and Finn}]{rafailov2023direct}
Rafael Rafailov, Archit Sharma, Eric Mitchell, Stefano Ermon, Christopher~D Manning, and Chelsea Finn. 2023.
\newblock Direct preference optimization: Your language model is secretly a reward model.
\newblock \emph{arXiv preprint arXiv:2305.18290}.

\bibitem[{Rei et~al.(2020)Rei, Stewart, Farinha, and Lavie}]{rei-etal-2020-comet}
Ricardo Rei, Craig Stewart, Ana~C Farinha, and Alon Lavie. 2020.
\newblock \href {https://doi.org/10.18653/v1/2020.emnlp-main.213} {{COMET}: A neural framework for {MT} evaluation}.
\newblock In \emph{Proceedings of the 2020 Conference on Empirical Methods in Natural Language Processing (EMNLP)}, pages 2685--2702, Online. Association for Computational Linguistics.

\bibitem[{Riley et~al.(2020)Riley, Caswell, Freitag, and Grangier}]{riley-etal-2020-translationese}
Parker Riley, Isaac Caswell, Markus Freitag, and David Grangier. 2020.
\newblock \href {https://doi.org/10.18653/v1/2020.acl-main.691} {Translationese as a language in {``}multilingual{''} {NMT}}.
\newblock In \emph{Proceedings of the 58th Annual Meeting of the Association for Computational Linguistics}, pages 7737--7746, Online. Association for Computational Linguistics.

\bibitem[{Robinson et~al.(2023)Robinson, Ogayo, Mortensen, and Neubig}]{robinson-etal-2023-chatgpt}
Nathaniel Robinson, Perez Ogayo, David~R. Mortensen, and Graham Neubig. 2023.
\newblock \href {https://doi.org/10.18653/v1/2023.wmt-1.40} {{C}hat{GPT} {MT}: Competitive for high- (but not low-) resource languages}.
\newblock In \emph{Proceedings of the Eighth Conference on Machine Translation}, pages 392--418, Singapore. Association for Computational Linguistics.

\bibitem[{Schulman et~al.(2017)Schulman, Wolski, Dhariwal, Radford, and Klimov}]{schulman2017proximal}
John Schulman, Filip Wolski, Prafulla Dhariwal, Alec Radford, and Oleg Klimov. 2017.
\newblock Proximal policy optimization algorithms.
\newblock \emph{arXiv preprint arXiv:1707.06347}.

\bibitem[{Sun et~al.(2023)Sun, Zhang, Deng, Cheng, and Huang}]{sun2023safety}
Hao Sun, Zhexin Zhang, Jiawen Deng, Jiale Cheng, and Minlie Huang. 2023.
\newblock Safety assessment of chinese large language models.
\newblock \emph{arXiv preprint arXiv:2304.10436}.

\bibitem[{Taori et~al.(2023)Taori, Gulrajani, Zhang, Dubois, Li, Guestrin, Liang, and Hashimoto}]{alpaca}
Rohan Taori, Ishaan Gulrajani, Tianyi Zhang, Yann Dubois, Xuechen Li, Carlos Guestrin, Percy Liang, and Tatsunori~B. Hashimoto. 2023.
\newblock Stanford alpaca: An instruction-following llama model.
\newblock \url{https://github.com/tatsu-lab/stanford_alpaca}.

\bibitem[{Touvron et~al.(2023)Touvron, Martin, Stone, Albert, Almahairi, Babaei, Bashlykov, Batra, Bhargava, Bhosale et~al.}]{touvron2023llama}
Hugo Touvron, Louis Martin, Kevin Stone, Peter Albert, Amjad Almahairi, Yasmine Babaei, Nikolay Bashlykov, Soumya Batra, Prajjwal Bhargava, Shruti Bhosale, et~al. 2023.
\newblock Llama 2: Open foundation and fine-tuned chat models.
\newblock \emph{arXiv preprint arXiv:2307.09288}.

\bibitem[{{\"U}st{\"u}n et~al.(2024){\"U}st{\"u}n, Aryabumi, Yong, Ko, D'souza, Onilude, Bhandari, Singh, Ooi, Kayid et~al.}]{ustun2024aya}
Ahmet {\"U}st{\"u}n, Viraat Aryabumi, Zheng-Xin Yong, Wei-Yin Ko, Daniel D'souza, Gbemileke Onilude, Neel Bhandari, Shivalika Singh, Hui-Lee Ooi, Amr Kayid, et~al. 2024.
\newblock Aya model: An instruction finetuned open-access multilingual language model.
\newblock \emph{arXiv preprint arXiv:2402.07827}.

\bibitem[{Wang et~al.(2023)Wang, Kordi, Mishra, Liu, Smith, Khashabi, and Hajishirzi}]{wang-etal-2023-self-instruct}
Yizhong Wang, Yeganeh Kordi, Swaroop Mishra, Alisa Liu, Noah~A. Smith, Daniel Khashabi, and Hannaneh Hajishirzi. 2023.
\newblock \href {https://doi.org/10.18653/v1/2023.acl-long.754} {Self-instruct: Aligning language models with self-generated instructions}.
\newblock In \emph{Proceedings of the 61st Annual Meeting of the Association for Computational Linguistics (Volume 1: Long Papers)}, pages 13484--13508, Toronto, Canada. Association for Computational Linguistics.

\bibitem[{Weissweiler et~al.(2023)Weissweiler, Hofmann, Kantharuban, Cai, Dutt, Hengle, Kabra, Kulkarni, Vijayakumar, Yu, Schuetze, Oflazer, and Mortensen}]{weissweiler-etal-2023-counting}
Leonie Weissweiler, Valentin Hofmann, Anjali Kantharuban, Anna Cai, Ritam Dutt, Amey Hengle, Anubha Kabra, Atharva Kulkarni, Abhishek Vijayakumar, Haofei Yu, Hinrich Schuetze, Kemal Oflazer, and David Mortensen. 2023.
\newblock \href {https://doi.org/10.18653/v1/2023.emnlp-main.401} {Counting the bugs in {C}hat{GPT}{'}s wugs: A multilingual investigation into the morphological capabilities of a large language model}.
\newblock In \emph{Proceedings of the 2023 Conference on Empirical Methods in Natural Language Processing}, pages 6508--6524, Singapore. Association for Computational Linguistics.

\bibitem[{Yang et~al.(2023)Yang, Li, Zhang, and Zong}]{yang2023bigtrans}
Wen Yang, Chong Li, Jiajun Zhang, and Chengqing Zong. 2023.
\newblock Bigtrans: Augmenting large language models with multilingual translation capability over 100 languages.
\newblock \emph{arXiv preprint arXiv:2305.18098}.

\bibitem[{Yang et~al.(2019)Yang, Zhang, Tar, and Baldridge}]{yang-etal-2019-paws}
Yinfei Yang, Yuan Zhang, Chris Tar, and Jason Baldridge. 2019.
\newblock \href {https://doi.org/10.18653/v1/D19-1382} {{PAWS}-{X}: A cross-lingual adversarial dataset for paraphrase identification}.
\newblock In \emph{Proceedings of the 2019 Conference on Empirical Methods in Natural Language Processing and the 9th International Joint Conference on Natural Language Processing (EMNLP-IJCNLP)}, pages 3687--3692, Hong Kong, China. Association for Computational Linguistics.

\bibitem[{Zhang et~al.(2020)Zhang, Williams, Titov, and Sennrich}]{zhang-etal-2020-improving}
Biao Zhang, Philip Williams, Ivan Titov, and Rico Sennrich. 2020.
\newblock \href {https://doi.org/10.18653/v1/2020.acl-main.148} {Improving massively multilingual neural machine translation and zero-shot translation}.
\newblock In \emph{Proceedings of the 58th Annual Meeting of the Association for Computational Linguistics}, pages 1628--1639, Online. Association for Computational Linguistics.

\bibitem[{Zhu et~al.(2023)Zhu, Lv, Dong, Yuan, Xu, Huang, Kong, Chen, and Li}]{zhu2023extrapolating}
Wenhao Zhu, Yunzhe Lv, Qingxiu Dong, Fei Yuan, Jingjing Xu, Shujian Huang, Lingpeng Kong, Jiajun Chen, and Lei Li. 2023.
\newblock Extrapolating large language models to non-english by aligning languages.
\newblock \emph{arXiv preprint arXiv:2308.04948}.

\end{thebibliography}

\clearpage
\newpage
\appendix

\section{Datasets}
\label{app:data}
We construct two datasets for multilingual tuning: multilingual instruction dataset in 100 languages and cross-lingual human feedback dataset in 30 languages.
We tokenize the inputs, instructions, and outputs of these two datasets separately, and present the statistics of the
results in Table~\ref{tab:data_token_sta}.
We observe that both LLaMA have a large average number of tokens, indicating that the original LLMs do not adequately support all languages in our constructed datasets. The sentences are segmented into smaller units, which increases the difficulty of capturing semantic information. Addressing the tokenizer issue in LLMs is an important step towards expanding LLMs to new languages, and we would like to address this in future work.
In addition, we show the languages covered in our constructed cross-lingual dataset in Table~\ref{tab:clhf_data_sta}.
Finally, the translator used for each language when translating the instructions and responses are shown in Table~\ref{tab:instruct_response_statistic}.

\begin{table}[!ht]
\centering
\begin{tabular}{ll|ll}
\toprule
\textbf{\#Langs}    & \textbf{Type} & \textbf{\#Langs}    & \textbf{Type} \\
\midrule
Arabic     & High & Russian    & High \\
Basque     & High & Slovak     & High \\
Bengali    & High & Spanish    & High \\
Chinese    & High & Swedish    & High \\
Croatian   & High & Ukrainian  & High \\
Danish     & High & Vietnamese & High \\
Dutch      & High & Armenian   & Low  \\
French     & High & Gujarati   & Low  \\
German     & High & Kannada    & Low  \\
Hindi      & High & Malayalam  & Low  \\
Hungarian  & High & Marathi    & Low  \\
Indonesian & High & Nepali     & Low  \\
Italian    & High & Serbian    & Low  \\
Portuguese & High & Tamil      & Low  \\
Romanian   & High & Telugu     & Low  \\
\bottomrule
\end{tabular}
\caption{\label{tab:clhf_data_sta}
Languages covered in our constructed cross-lingual human feedback dataset.
}
\end{table}

\section{Model Configuration}
\label{app:hyper_para}
The training process of \md contains two steps.
In supervised fine-tuning, we use LORA to fine-tune LLMs on our constructed multilingual instruction dataset.
In aligning LLMs with human feedback, we use DPO with LORA to fine-tune the LLMs.
We train \md on one machine with 8 A100 80GB GPUs.
The hyper-parameters for \md are presented in Table~\ref{tab:hyper_para}.

\begin{table}[!ht]
\centering
\resizebox{\columnwidth}{!}{
\begin{tabular}{lc|lc|lc}
\toprule
\multicolumn{2}{c}{\textbf{SFT}} & \multicolumn{2}{c}{\textbf{LORA}} & \multicolumn{2}{c}{\textbf{DPO}} \\
\midrule
batch size      & 4     & r            & 8         & batch size      & 8     \\
epoch           & 1     & alpha        & 16        & epoch           & 1     \\
learning rate   & 1e-4  & dropout      & 0.05      & learning rate   & 5e-4  \\
max length      & 1024  &              &           & max length      & 1024  \\
\bottomrule
\end{tabular}}
\caption{\label{tab:hyper_para}
Hyper-parameters for multilingual tuning.
}
\end{table}
\vspace{-1em}

\section{Prompt Setting}
\label{app:prompt}
To evaluate the generating capability of LLMs, we use English as the prompt language as shown in Table~\ref{tab:prompt}.
To evaluate the understanding capability of LLMs, we use Google Translate API to translate the English prompt to the other languages.

\section{Complete Results}
\label{app:full_res}
We present the results of all languages for each benchmark: PAWS-X (Table~\ref{tab:paws_x_res}), XCOPA (Table~\ref{tab:xcopa_res}), Self-Instruct* (Table~\ref{tab:self_res}), XL-Sum (Table~\ref{tab:xl_sum_gen_res}, Table~\ref{tab:xl_sum_und_res}) and FLORES (Table~\ref{tab:gen_flores_from_1},~\ref{tab:gen_flores_from_2},~\ref{tab:gen_flores_to_1},~\ref{tab:gen_flores_to_2},~\ref{tab:und_flores_from_1},~\ref{tab:und_flores_from_2},~\ref{tab:und_flores_to_1} and Table~\ref{tab:und_flores_to_2}).

\begin{table}[!ht]
\small
\centering
\begin{tabular}{@{}l@{\hspace{5pt}}ccccccc@{}}
\toprule
\multicolumn{8}{c}{Language of inputs and outputs} \\
  \textbf{LLM}      & \textbf{En}   & \textbf{Zh}   & \textbf{Vi}   & \textbf{Tr}   & \textbf{Ar}   & \textbf{El}   & \textbf{Hi}    \\
\midrule
\multicolumn{8}{@{}l}{\emph{Understanding capability (Instruction identical to inputs)}} \\
ChatGPT & \textbf{56.0} & 20.5 & 26.8 & 18.3 & 24.1 & 17.7 & 0.6 \\
LLaMA & \textbf{76.6} & 27.2 & 36.6 & 27.8 & 11.8 & 22.3 & 14.3 \\
BLOOM  & \textbf{83.9} & 83.0 & 79.9 & 27.4 & 79.2 & 22.8 & 82.7   \\
\midrule
\multicolumn{8}{@{}l}{\emph{Generation capability (Instructions in English)}} \\
ChatGPT & \textbf{56.0} & 37.1 & 36.1 & 34.5 & 32.0 & 29.7 & 17.5 \\
LLaMA  & \textbf{76.6} & 66.3 & 42.9 & 38.1 & 24.2 & 40.7 & 30.8 \\
BLOOM   & \textbf{83.9} & 81.8 & 79.2 & 27.6 & 77.2 & 49.2 & 80.8  \\
\bottomrule
\end{tabular}
\caption{\label{tab:intro_multi_cap}
A primary evaluation for the multilingual capability (understanding and generation) of LLMs (ChatGPT, LLaMA and BLOOM) on the XQuAD dataset~\cite{artetxe-etal-2020-cross} in terms of exact match (EM). 
To evaluate the \textit{understanding capability} we use an identical language to represent instructions and inputs to LLMs. 
To evaluate the \textit{generating capability} instructions are always in English, regardless of the language of input text.
}
\end{table}

\section{Multilingual Capability of LLMs}
\label{app:multi_intro}
Table~\ref{tab:intro_multi_cap} illustrates the significant variation in performance achieved by LLMs across different examined languages, with the highest scores observed when tasks are described and presented in English.
Furthermore, LLMs exhibit varying degrees of understanding capabilities across different languages, with some cases where they fail to understand the language.
For instance, when using \textit{Hindi} as the instruction language, ChatGPT lacks an understanding of the instructions, leading to subpar performance.

\begin{table*}[!ht]
\centering
\resizebox{\textwidth}{!}{
\begin{tabular}{l|l|l||l|l|l}
\toprule
\textbf{\#Lang.} & \textbf{Ins\_trans} & \textbf{Response} & \textbf{\#Lang.} & \textbf{Ins\_trans} & \textbf{Response} \\
\midrule
afrikaans     & Google Translate & ChatGPT          & norwegian      & Google Translate & ChatGPT          \\
albanian      & Google Translate & ChatGPT          & persian        & Google Translate & ChatGPT          \\
amharic       & Google Translate & Google Translate & polish         & Google Translate & ChatGPT          \\
arabic        & Google Translate & ChatGPT          & portuguese     & Google Translate & ChatGPT          \\
armenian      & Google Translate & Google Translate & romanian       & Google Translate & ChatGPT          \\
azerbaijani   & Google Translate & Google Translate & russian        & Google Translate & ChatGPT          \\
belarusian    & Google Translate & Google Translate & serbian        & Google Translate & Google Translate \\
bengali       & Google Translate & Google Translate & sindhi         & Google Translate & Google Translate \\
bosnian       & Google Translate & ChatGPT          & sinhala        & Google Translate & Google Translate \\
bulgarian     & Google Translate & ChatGPT          & slovak         & Google Translate & ChatGPT          \\
catalan       & Google Translate & ChatGPT          & slovenian      & Google Translate & ChatGPT          \\
cebuano       & Google Translate & ChatGPT          & somali         & Google Translate & Google Translate \\
chinese       & Google Translate & ChatGPT          & spanish        & Google Translate & ChatGPT          \\
croatian      & Google Translate & ChatGPT          & sundanese      & Google Translate & Google Translate \\
czech         & Google Translate & ChatGPT          & swahili        & Google Translate & ChatGPT          \\
danish        & Google Translate & ChatGPT          & swedish        & Google Translate & ChatGPT          \\
dutch         & Google Translate & ChatGPT          & tamil          & Google Translate & Google Translate \\
english       & Google Translate & ChatGPT          & thai           & Google Translate & Google Translate \\
estonian      & Google Translate & ChatGPT          & turkish        & Google Translate & Google Translate \\
finnish       & Google Translate & ChatGPT          & ukrainian      & Google Translate & ChatGPT          \\
french        & Google Translate & ChatGPT          & urdu           & Google Translate & ChatGPT          \\
galician      & Google Translate & ChatGPT          & uzbek          & Google Translate & Google Translate \\
georgian      & Google Translate & Google Translate & vietnamese     & Google Translate & ChatGPT          \\
german        & Google Translate & ChatGPT          & welsh          & Google Translate & ChatGPT          \\
gujarati      & Google Translate & Google Translate & xhosa          & Google Translate & Google Translate \\
hausa         & Google Translate & Google Translate & yiddish        & Google Translate & Google Translate \\
hebrew        & Google Translate & ChatGPT          & yoruba         & Google Translate & Google Translate \\
hindi         & Google Translate & ChatGPT          & zulu           & Google Translate & Google Translate \\
hungarian     & Google Translate & ChatGPT          & asturian       & NLLB             & ChatGPT          \\
icelandic     & Google Translate & ChatGPT          & bashkir        & NLLB             & NLLB             \\
igbo          & Google Translate & Google Translate & breton         & NLLB             & NLLB             \\
indonesian    & Google Translate & ChatGPT          & burmese        & NLLB             & NLLB             \\
irish         & Google Translate & ChatGPT          & frisian        & NLLB             & NLLB             \\
italian       & Google Translate & ChatGPT          & fulah          & NLLB             & NLLB             \\
japanese      & Google Translate & ChatGPT          & gaelic         & NLLB             & NLLB             \\
javanese      & Google Translate & ChatGPT          & ganda          & NLLB             & NLLB             \\
kannada       & Google Translate & Google Translate & greek          & NLLB             & ChatGPT          \\
kazakh        & Google Translate & Google Translate & haitian        & NLLB             & ChatGPT          \\
korean        & Google Translate & ChatGPT          & iloko          & NLLB             & NLLB             \\
lao           & Google Translate & Google Translate & khmer          & NLLB             & NLLB             \\
latvian       & Google Translate & ChatGPT          & lingala        & NLLB             & NLLB             \\
lithuanian    & Google Translate & ChatGPT          & northern sotho & NLLB             & NLLB             \\
luxembourgish & Google Translate & ChatGPT          & occitan        & NLLB             & ChatGPT          \\
macedonian    & Google Translate & ChatGPT          & oriya          & NLLB             & NLLB             \\
malagasy      & Google Translate & Google Translate & panjabi        & NLLB             & NLLB             \\
malay         & Google Translate & ChatGPT          & pashto         & NLLB             & NLLB             \\
malayalam     & Google Translate & Google Translate & swati          & NLLB             & NLLB             \\
marathi       & Google Translate & Google Translate & tagalog        & NLLB             & ChatGPT          \\
mongolian     & Google Translate & Google Translate & tswana         & NLLB             & NLLB             \\
nepali        & Google Translate & Google Translate & wolof          & NLLB             & NLLB             \\
\bottomrule
\end{tabular}
}
\caption{\label{tab:instruct_response_statistic}
The translator used for each language when translating the instruction described in Section~\ref{sec:method_mid}.
}
\end{table*}
\begin{table*}[!ht]
\centering
\resizebox{\textwidth}{!}{
\begin{tabular}{lccccc}
\toprule
\multicolumn{6}{c}{\textbf{Multilingual Instruction Datasets}}  \\
\midrule
Tokenizer & Vocab size & Instruction tokens & Input tokens & Response (acc) tokens & Response (rej) tokens \\
m2m\_100  & 128,104    & 19.25              & 37.85        & 180.35                & -                      \\
LLaMA     & 32,000     & 38.49              & 65.16        & 370.44                & -                     \\
BLOOM     & 251,680    & 25.20              & 44.62        & 239.76                & -                     \\
\midrule
\multicolumn{6}{c}{\textbf{Cross-Lingual Human Feedback Datasets}}            \\
\midrule
m2m\_100  & 128,104    & 20.31              & 20.54        & 100.96                & 101.2                 \\
LLaMA     & 32,000     & 52.32              & 49.27        & 259.08                & 260.61                \\
BLOOM     & 251,680    & 20.08              & 20.17        & 98.38                 & 99.24                 \\
\bottomrule
\end{tabular}}
\caption{\label{tab:data_token_sta}
Statistics of average tokens in each instructions, inputs and outputs on our constructed datasets.
}
\end{table*}

\section{Comparison with Fine-tuned SLM}
\label{app:slms}
While it is not feasible to evaluate a single small language model (SLM) across all the benchmarks we employed, we omit these results from Table~\ref{tab:main_res}.
However, we do provide some experimental findings for reference.
As shown in Table~\ref{tab:slms_xlsum}, our model does not perform as well as certain SLM fine-tuned on specific benchmarks (e.g., XL-Sum), we successfully narrow the performance gap between the original large language models (e.g., LLaMA) and closed-source large models (e.g., ChatGPT).
This achievement is particularly meaningful as we refrain from using any task-specific data for fine-tuning.
Interestingly, when SLM are not fine-tuned for a specific task, Table~\ref{tab:slms_self_instruct} demonstrates that our model significantly outperform the SLM in the Self-Instruct* benchmark.

\begin{table*}[!ht]
\centering
\begin{tabular}{lp{0.75\textwidth}}
\toprule
\textbf{Benchmark}     & \textbf{Prompt}   \\
\midrule
PAWS-X        & The following are two \{\texttt{lang\_name}\} sentences. Sentence 1: \{\texttt{sentence1}\} Sentence 2: \{\texttt{sentence2}\} Does sentence 1 paraphrase sentence 2? yes or no?                                               \\
\midrule
XCOPA         & Here is a premise: \{\texttt{premise}\}. What is the \{\texttt{question}\}? Help me to pick the more plausible option -A: \{\texttt{choice1}\}, -B: \{\texttt{choice2}\}                                                                \\
\midrule
Self-Instruct & Instruction: \{\texttt{instruction}\}                                                                                                                                                                        \\
\midrule
XL-Sum        & The following is an article. Article: \{\texttt{article}\} Please summarize the provided article.                                                                                                            \\
\midrule
FLORES        & Translate the following \{\texttt{source\_lang}\} sentence from \{\texttt{source\_lang}\} to \{\texttt{target\_lang}\}. \{\texttt{souce\_lang}\} Sentence: \{\texttt{source\_sentence}\} Please only return the translated \{\texttt{target\_lang}\} text \\
\bottomrule
\end{tabular}
\caption{\label{tab:prompt}
The prompt of each benchmark used in our experiments.
}
\end{table*}

\begin{table*}[!ht]
\resizebox{\textwidth}{!}{
\begin{tabular}{lccccccccc}
\toprule
          & \multicolumn{3}{c}{\textbf{Low}}     & \multicolumn{3}{c}{\textbf{Medium}}  & \multicolumn{3}{c}{\textbf{High}}    \\
\cmidrule(lr){2-4}\cmidrule(lr){5-7}\cmidrule(lr){8-10}
          & ROUGE-1 & ROUGE-2 & ROUGE-L & ROUGE-1 & ROUGE-2 & ROUGE-L & ROUGE-1 & ROUGE-2 & ROUGE-L \\
\midrule
mT5-XXL   & 24.89   & 9.75    & 21.08   & 31.20   & 12.91   & 25.50   & 33.14   & 14.44   & 27.57   \\
chatGPT   & 15.85   & 4.78    & 11.24   & 18.68   & 5.68    & 12.99   & 19.15   & 5.80    & 13.65   \\
LLaMA     & 6.26    & 2.47    & 6.52    & 5.80    & 2.18    & 7.27    & 8.08    & 2.59    & 7.36    \\
xLLMs-100 & 9.99    & 3.75    & 9.27    & 13.57   & 5.03    & 11.28   & 16.61   & 5.27    & 11.33   \\
\bottomrule
\end{tabular}}
\caption{
\label{tab:slms_xlsum}
Compare with mT5-XXL finetuned model in XL-Sum benchmark.
}
\end{table*}

\begin{table*}[!ht]
\centering
\begin{tabular}{lcccccc}
\toprule
          & \multicolumn{3}{c}{\textbf{Low}}     & \multicolumn{3}{c}{\textbf{High}}    \\
\cmidrule(lr){2-4}\cmidrule(lr){5-7}
          & ROUGE-1 & ROUGE-2 & ROUGE-L & ROUGE-1 & ROUGE-2 & ROUGE-L \\
\midrule
T5-LM     & 1.23    & 0.35    & 1.86    & 3.17    & 0.83    & 2.67    \\
chatGPT   & 21.93   & 8.24    & 15.23   & 31.60   & 13.81   & 25.15   \\
LLaMA     & 5.38    & 2.16    & 4.37    & 8.81    & 2.15    & 6.24    \\
xLLMs-100 & 9.16    & 5.25    & 9.33    & 14.71   & 7.28    & 12.75   \\
\bottomrule
\end{tabular}
\caption{
\label{tab:slms_self_instruct}
Compare with T5-LM in Self-Instruct benchmark.
}
\end{table*}

\begin{table*}[!ht]
\centering
\begin{tabular}{lcccccccc}
\toprule
\multicolumn{9}{c}{\textbf{Generating Capability}} \\
\midrule
           & de    & en    & es    & fr    & ja    & ko    & zh    & Avg.  \\
\midrule
LLaMA      & 45.65 & 62.91 & 50.06 & 46.55 & 46.40 & 48.00 & 51.98 & 50.22 \\
\bxllama   & 44.50 & 63.43 & 46.43 & 45.15 & 45.05 & 46.40 & 47.88 & 48.41 \\
\sftllama  & 47.80 & 65.44 & 48.32 & 48.20 & 47.15 & 47.75 & 47.85 & 50.36 \\
\md        & \textbf{55.25} & \textbf{70.68} & \textbf{70.68} & \textbf{54.85} & \textbf{55.70} & \textbf{55.15} & \textbf{71.25} & \textbf{61.94} \\
\midrule
BLOOM      & 45.00 & 62.35 & 45.45 & 45.15 & 44.10 & 44.80 & 44.85 & 47.39 \\
\bxbloom   & 44.50 & 62.49 & 45.43 & 45.25 & 43.95 & 44.40 & 44.81 & 47.26 \\
\sftbloom  & 46.50 & 63.44 & 46.18 & 46.10 & 45.45 & 45.85 & 46.00 & 48.50 \\
\md        & \textbf{47.80} & \textbf{65.21} & \textbf{49.49} & \textbf{48.10} & \textbf{47.10} & \textbf{47.70} & \textbf{48.29} & \textbf{50.53} \\
\midrule
\multicolumn{9}{c}{\textbf{Understanding Capability}} \\
\midrule
           & de    & en    & es    & fr    & ja    & ko    & zh    & Avg.  \\
\midrule
LLaMA      & 32.61 & 62.91 & 42.43 & 33.43 & 33.53 & 30.93 & 30.86 & 38.10 \\
\bxllama   & 34.23 & 63.43 & 33.70 & 31.68 & \textbf{35.73} & 30.93 & \textbf{31.24} & 37.28 \\
\sftllama  & \textbf{34.51} & 65.44 & 69.80 & 31.20 & 33.47 & 30.94 & 30.89 & 42.32 \\
\md        & 31.27 & \textbf{70.68} & \textbf{70.68} & \textbf{63.49} & 30.72 & \textbf{30.94} & 30.89 & \textbf{46.95} \\
\midrule
BLOOM      & 35.00 & 62.35 & \textbf{31.84} & 31.77 & 32.59 & 30.86 & 30.89 & 36.47 \\
\bxbloom   & 34.83 & 62.49 & 31.20 & 31.28 & 32.79 & 31.20 & \textbf{31.12} & 36.42 \\
\sftbloom  & \textbf{35.13} & 63.44 & 31.29 & 32.54 & 32.41 & 30.99 & 30.92 & 36.67 \\
\md        & 34.08 & \textbf{65.21} & 31.20 & \textbf{49.69} & \textbf{33.84} & \textbf{33.86} & 30.92 & \textbf{39.83} \\
\bottomrule
\end{tabular}
\caption{\label{tab:paws_x_res}
Accuracy scores on the PAWS-X benchmark~\cite{yang-etal-2019-paws}.
}
\end{table*}
\begin{table*}[!ht]
\centering
\resizebox{\textwidth}{!}{
\begin{tabular}{lrrrrrrrrrrrr}
\toprule
\multicolumn{13}{c}{\textbf{Generating Capability}} \\
\midrule
           & ar    & cs    & de    & zh    & hi    & hy    & ky   & yo    & ta    & mn    & Avg\_L & Avg\_H \\
\midrule
LLaMA      & 7.09  & 7.95  & 8.70  & 15.23 & 5.10  & 5.39  & 4.85 & 6.95  & 5.44  & 4.28  & 5.38    & 8.81    \\
\bxllama   & 9.11  & 7.57  & 9.51  & 14.27 & 8.53  & 5.52  & 6.94 & 10.43 & 5.79  & 6.37  & 7.01    & 9.80    \\
\sftllama  & 10.06 & 8.54  & 9.77  & 19.40 & 12.98 & 7.77  & 5.27 & 9.73  & 7.62  & 5.11  & 7.10    & 12.15   \\
\md        & 12.44 & 12.78 & 15.49 & 19.90 & 12.94 & 3.79  & 9.28 & 11.48 & 10.82 & 10.42 & \textbf{9.16}    & \textbf{14.71}   \\
\midrule
BLOOM      & 6.52  & 4.90  & 6.12  & 11.13 & 6.41  & 2.37  & 3.44 & 5.99  & 5.20  & 3.34  & 4.07    & 7.01    \\
\bxbloom   & 5.74  & 7.08  & 9.40  & 12.26 & 6.55  & 3.32  & 6.58 & 8.15  & 5.34  & 6.02  & 5.88    & 8.21    \\
\bottomrule
\end{tabular}}
\caption{\label{tab:self_res}
ROUGE-1 scores on the Self-Instruct* benchmark~\cite{wang-etal-2023-self-instruct} and average scores on low-resource (Avg\_L) and high-resource (Avg\_H) languages.
}
\end{table*}
\begin{table*}[!ht]
\centering
\resizebox{\textwidth}{!}{
\begin{tabular}{lccccccccccccc}
\toprule
\multicolumn{14}{c}{\textbf{Generating Capability}} \\
\midrule
           & et    & ht    & id    & it    & qu    & sw    & ta    & th    & tr    & vi    & zh    & Avg\_L & Avg\_H \\
\midrule
LLaMA      & 50.00 & 49.80 & 51.00 & 51.80 & 48.40 & 50.40 & 49.80 & 52.40 & 50.60 & 50.80 & 55.20 & 49.33   & 51.52   \\
\bxllama   & 50.60 & 47.60 & 49.40 & 49.40 & 48.60 & 50.20 & 47.80 & 49.20 & 49.00 & 50.60 & 50.40 & 48.00   & 49.85   \\
\sftllama  & 50.40 & 50.00 & 49.60 & 50.00 & 47.00 & 50.00 & 49.80 & 50.00 & 50.20 & 50.20 & 50.00 & 48.93   & 50.05   \\
\md        & 50.00 & 51.00 & 54.80 & 60.60 & 48.20 & 52.60 & 49.93 & 53.20 & 54.00 & 51.20 & 61.00 & 49.71   & \textbf{54.68}   \\
\midrule
BLOOM      & 48.00 & 49.16 & 47.84 & 48.27 & 50.22 & 50.63 & 50.16 & 50.83 & 50.72 & 49.37 & 50.11 & 49.85   & 49.47   \\
\bxbloom   & 48.14 & 43.57 & 50.60 & 49.32 & 49.80 & 50.20 & 49.80 & 50.60 & 50.40 & 48.88 & 51.67 & 47.72   & 49.98   \\
\bottomrule
\end{tabular}}
\caption{\label{tab:xcopa_res}
Accuracy scores on the XCOPA benchmark~\cite{ponti-etal-2020-xcopa}.
}
\end{table*}
\begin{table*}[!ht]
\centering
\resizebox{\textwidth}{!}{
\begin{tabular}{ll|rrrr|rrrr}
\toprule
\#Langs.             & Type   & LLaMA & \bxllama & \sftllama & \md    & BLOOM & \bxbloom & \sftbloom & \md    \\
\midrule
marathi              & High   & 0.64  & 0.01    & 4.00     & 5.77  & 0.90  & 1.97    & 3.12     & 6.48  \\
english              & High   & 16.43 & 19.06   & 18.29    & 20.77 & 12.66 & 15.75   & 12.19    & 16.91 \\
pashto               & High   & 17.17 & 0.67    & 17.21    & 19.02 & 22.03 & 8.26    & 7.96     & 21.20 \\
thai                 & Medium & 15.55 & 8.35    & 15.69    & 18.26 & 9.47  & 5.02    & 8.89     & 10.37 \\
ukrainian            & High   & 1.24  & 0.50    & 1.25     & 10.93 & 4.87  & 2.05    & 6.77     & 8.13  \\
serbian\_cyrillic    & Medium & 1.73  & 0.47    & 0.56     & 9.65  & 3.76  & 1.40    & 12.23    & 10.33 \\
spanish              & High   & 6.37  & 3.86    & 4.40     & 21.43 & 14.76 & 10.49   & 23.99    & 10.70 \\
\bottomrule
\end{tabular}}
\caption{\label{tab:xl_sum_gen_res}
\textbf{Generating Capability:} ROUGE-1 scores on the XL-Sum benchmark~\cite{hasan-etal-2021-xl} and average scores on low-resource (Avg\_L), medium-resource (Avg\_M) and high-resource (Avg\_H) languages.
}
\end{table*}

\begin{table*}[!ht]
\centering
\resizebox{\textwidth}{!}{
\begin{tabular}{ll|rrrr|rrrr}
\toprule
\#Langs.             & Type   & LLaMA & \bxllama & \sftllama & \md    & BLOOM & \bxbloom & \sftbloom & \md    \\
\midrule
marathi              & High   & 0.70  & 1.55    & 4.65     & 4.99  & 10.45 & 7.84    & 8.01     & 12.43 \\
english              & High   & 20.06 & 19.07   & 18.28    & 18.74 & 13.37 & 15.78   & 12.19    & 18.65 \\
pashto               & High   & 5.81  & 12.74   & 9.92     & 21.49 & 18.72 & 0.02    & 7.86     & 20.76 \\
thai                 & Medium & 7.92  & 11.48   & 13.48    & 18.65 & 13.47 & 2.90    & 7.69     & 11.37 \\
ukrainian            & High   & 0.69  & 0.88    & 2.23     & 12.10 & 9.06  & 4.21    & 7.61     & 10.61 \\
serbian\_cyrillic    & Medium & 0.66  & 4.96    & 0.84     & 5.68  & 9.81  & 4.07    & 8.63     & 11.75 \\
spanish              & High   & 7.77  & 17.11   & 4.99     & 22.43 & 19.42 & 20.14   & 26.30    & 21.47 \\
\bottomrule
\end{tabular}}
\caption{\label{tab:xl_sum_und_res}
\textbf{Understanding Capability:} ROUGE-1 scores on the XL-Sum benchmark~\cite{hasan-etal-2021-xl} and average scores on low-resource (Avg\_L), medium-resource (Avg\_M) and high-resource (Avg\_H) languages.
}
\end{table*}
\begin{table*}[t]
\centering
\resizebox{\textwidth}{!}{
\begin{tabular}{ll|rrrr|rrrr}
\toprule
\#Langs.             & Type   & LLaMA & \bxllama & \sftllama & \md    & BLOOM & \bxbloom & \sftbloom & \md    \\
\midrule
amh\_Ethi & Low  & 0.33  & 0.11    & 0.52     & 0.26  & 0.18  & 0.13     & 0.62     & 2.82  \\
arb\_Arab & High & 2.14  & 2.65    & 2.25     & 3.85  & 0.28  & 0.40     & 2.25     & 5.29  \\
asm\_Beng & Low  & 0.30  & 0.97    & 0.59     & 0.63  & 0.35  & 0.10     & 1.30     & 3.59  \\
ast\_Latn & Low  & 2.70  & 3.13    & 5.56     & 6.76  & 1.79  & 1.26     & 5.57     & 5.69  \\
azj\_Latn & Low  & 1.36  & 1.26    & 2.37     & 2.41  & 0.38  & 0.39     & 2.30     & 3.71  \\
bel\_Cyrl & Low  & 1.60  & 1.71    & 1.95     & 2.81  & 0.61  & 0.31     & 1.92     & 3.94  \\
ben\_Beng & High & 1.03  & 0.94    & 1.00     & 1.06  & 0.37  & 0.11     & 1.34     & 5.09  \\
\bottomrule
\end{tabular}}
\caption{\label{tab:gen_flores_from_1}
\textbf{Generating Capability (Part I):} BLEU scores on the FLORES benchmark~\cite{goyal-etal-2022-flores} and average scores on low-resource (Avg\_L) and high-resource (Avg\_H) languages. We report on translating from English to into other languages.
}
\end{table*}

\begin{table*}[t]
\centering
\resizebox{\textwidth}{!}{
\begin{tabular}{ll|rrrr|rrrr}
\toprule
\#Langs.             & Type   & LLaMA & \bxllama & \sftllama & \md    & BLOOM & \bxbloom & \sftbloom & \md    \\
\midrule
luo\_Latn & Low  & 1.27  & 0.61    & 2.73     & 2.43  & 0.95  & 0.45     & 2.12     & 3.93  \\
lvs\_Latn & High & 1.56  & 2.00    & 2.62     & 3.33  & 0.55  & 0.42     & 2.55     & 3.96  \\
mal\_Mlym & Low  & 0.68  & 0.71    & 1.94     & 1.48  & 0.64  & 0.53     & 2.41     & 3.58  \\
mar\_Deva & Low  & 1.20  & 1.47    & 1.63     & 2.06  & 0.59  & 0.33     & 2.22     & 4.09  \\
mkd\_Cyrl & High & 2.16  & 3.09    & 2.54     & 5.43  & 0.70  & 0.13     & 2.02     & 4.11  \\
mlt\_Latn & High & 1.85  & 1.81    & 3.34     & 3.82  & 0.98  & 0.57     & 3.40     & 3.65  \\
khk\_Cyrl & Low  & 0.47  & 0.02    & 2.16     & 1.61  & 0.46  & 0.25     & 1.99     & 3.63  \\
\bottomrule
\end{tabular}}
\caption{\label{tab:gen_flores_from_2}
\textbf{Generating Capability (Part II):} BLEU scores on the FLORES benchmark~\cite{goyal-etal-2022-flores} and average scores on low-resource (Avg\_L) and high-resource (Avg\_H) languages. We report on translating from English to into other languages.
}
\end{table*}

\begin{table*}[t]
\centering
\resizebox{\textwidth}{!}{
\begin{tabular}{ll|rrrr|rrrr}
\toprule
\#Langs.             & Type   & LLaMA & \bxllama & \sftllama & \md    & BLOOM & \bxbloom & \sftbloom & \md    \\
\midrule
amh\_Ethi & Low  & 1.00  & 0.57    & 1.29     & 1.83  & 0.20  & 0.34     & 1.84     & 0.66  \\
arb\_Arab & High & 3.19  & 1.08    & 3.16     & 10.80 & 2.89  & 3.37     & 3.18     & 14.78 \\
asm\_Beng & Low  & 1.41  & 0.34    & 1.17     & 2.29  & 0.62  & 2.89     & 2.96     & 5.32  \\
ast\_Latn & Low  & 5.38  & 2.17    & 11.02    & 21.11 & 1.46  & 3.64     & 4.98     & 15.45 \\
azj\_Latn & Low  & 1.71  & 1.86    & 1.88     & 3.79  & 1.14  & 1.76     & 1.19     & 2.32  \\
bel\_Cyrl & Low  & 2.39  & 1.53    & 1.87     & 6.32  & 1.18  & 1.72     & 1.51     & 1.71  \\
ben\_Beng & High & 1.65  & 0.44    & 1.23     & 3.13  & 0.89  & 2.28     & 3.56     & 12.00 \\
\bottomrule
\end{tabular}}
\caption{\label{tab:gen_flores_to_1}
\textbf{Generating Capability (Part I):} BLEU scores on the FLORES benchmark~\cite{goyal-etal-2022-flores} and average scores on low-resource (Avg\_L) and high-resource (Avg\_H) languages. We report on translating from the other languages into English.
}
\end{table*}

\begin{table*}[t]
\centering
\resizebox{\textwidth}{!}{
\begin{tabular}{ll|rrrr|rrrr}
\toprule
\#Langs.             & Type   & LLaMA & \bxllama & \sftllama & \md    & BLOOM & \bxbloom & \sftbloom & \md    \\
\midrule
luo\_Latn & Low  & 1.37  & 1.42    & 1.59     & 2.64  & 1.21  & 2.32     & 1.32     & 2.16  \\
lvs\_Latn & High & 1.90  & 1.67    & 3.67     & 4.36  & 0.91  & 1.81     & 1.34     & 3.31  \\
mal\_Mlym & Low  & 1.40  & 1.04    & 1.20     & 2.10  & 0.62  & 2.68     & 2.94     & 6.26  \\
mar\_Deva & Low  & 1.57  & 1.18    & 1.18     & 3.09  & 0.99  & 2.33     & 2.64     & 10.29 \\
mkd\_Cyrl & High & 4.91  & 1.61    & 10.99    & 18.43 & 0.89  & 1.62     & 2.24     & 3.59  \\
mlt\_Latn & High & 2.20  & 2.55    & 2.95     & 6.69  & 1.56  & 2.86     & 2.05     & 2.60  \\
khk\_Cyrl & Low  & 1.28  & 1.61    & 1.77     & 2.27  & 1.04  & 1.71     & 1.31     & 1.84  \\
\bottomrule
\end{tabular}}
\caption{\label{tab:gen_flores_to_2}
\textbf{Generating Capability (Part II):} BLEU scores on the FLORES benchmark~\cite{goyal-etal-2022-flores} and average scores on low-resource (Avg\_L) and high-resource (Avg\_H) languages. We report on translating from the other languages into English.
}
\end{table*}

\begin{table*}[t]
\centering
\resizebox{\textwidth}{!}{
\begin{tabular}{ll|rrrr|rrrr}
\toprule
\#Langs.             & Type   & LLaMA & \bxllama & \sftllama & \md    & BLOOM & \bxbloom & \sftbloom & \md    \\
\midrule
amh\_Ethi & Low  & 0.22  & 0.52    & 0.28     & 0.26  & 0.41  & 0.20     & 0.44     & 0.54  \\
arb\_Arab & High & 2.29  & 1.44    & 2.31     & 2.45  & 1.33  & 2.02     & 5.09     & 6.10  \\
asm\_Beng & Low  & 0.39  & 9.90    & 7.68     & 6.76  & 2.31  & 3.51     & 2.26     & 0.72  \\
ast\_Latn & Low  & 5.54  & 5.10    & 8.76     & 1.82  & 6.09  & 4.66     & 7.12     & 1.94  \\
azj\_Latn & Low  & 5.44  & 6.17    & 4.25     & 6.15  & 9.03  & 6.62     & 1.15     & 9.84  \\
bel\_Cyrl & Low  & 1.35  & 1.48    & 1.31     & 1.56  & 1.20  & 1.41     & 1.83     & 1.47  \\
ben\_Beng & High & 0.88  & 0.98    & 0.93     & 0.75  & 1.30  & 0.84     & 2.98     & 3.15  \\
\bottomrule
\end{tabular}}
\caption{\label{tab:und_flores_from_1}
\textbf{Understanding Capability (Part I):} BLEU scores on the FLORES benchmark~\cite{goyal-etal-2022-flores} and average scores on low-resource (Avg\_L) and high-resource (Avg\_H) languages. We report on translating from English into the other languages.
}
\end{table*}

\begin{table*}[t]
\centering
\resizebox{\textwidth}{!}{
\begin{tabular}{ll|rrrr|rrrr}
\toprule
\#Langs.             & Type   & LLaMA & \bxllama & \sftllama & \md    & BLOOM & \bxbloom & \sftbloom & \md    \\
\midrule
luo\_Latn & Low  & 0.83  & 2.48    & 8.55     & 2.32  & 3.56  & 2.09     & 3.15     & 6.90  \\
lvs\_Latn & High & 1.94  & 1.69    & 1.49     & 2.14  & 1.17  & 0.94     & 2.14     & 1.61  \\
mal\_Mlym & Low  & 0.28  & 2.07    & 1.74     & 1.32  & 0.90  & 0.77     & 1.86     & 1.42  \\
mar\_Deva & Low  & 1.25  & 1.59    & 1.59     & 1.57  & 0.98  & 0.69     & 2.12     & 3.01  \\
mkd\_Cyrl & High & 2.38  & 2.01    & 2.07     & 5.07  & 1.15  & 1.12     & 2.18     & 1.83  \\
mlt\_Latn & High & 1.63  & 2.48    & 1.41     & 3.03  & 1.76  & 1.11     & 2.15     & 1.80  \\
khk\_Cyrl & Low  & 1.60  & 1.21    & 0.79     & 0.99  & 1.22  & 0.96     & 1.83     & 1.14  \\
\bottomrule
\end{tabular}}
\caption{\label{tab:und_flores_from_2}
\textbf{Understanding Capability (Part II):} BLEU scores on the FLORES benchmark~\cite{goyal-etal-2022-flores} and average scores on low-resource (Avg\_L) and high-resource (Avg\_H) languages. We report on translating from English into the other languages.
}
\end{table*}

\begin{table*}[t]
\centering
\resizebox{\textwidth}{!}{
\begin{tabular}{ll|rrrr|rrrr}
\toprule
\#Langs.             & Type   & LLaMA & \bxllama & \sftllama & \md    & BLOOM & \bxbloom & \sftbloom & \md    \\
\midrule
amh\_Ethi & Low  & 0.15  & 0.14    & 1.22     & 1.60  & 1.84  & 0.12     & 0.24     & 1.33  \\
arb\_Arab & High & 2.11  & 2.85    & 6.54     & 6.83  & 3.18  & 3.00     & 4.05     & 16.75 \\
asm\_Beng & Low  & 3.26  & 5.08    & 6.03     & 1.76  & 2.96  & 2.75     & 0.35     & 0.63  \\
ast\_Latn & Low  & 2.90  & 6.78    & 1.15     & 2.12  & 4.98  & 4.19     & 3.48     & 2.60  \\
azj\_Latn & Low  & 1.99  & 8.30    & 1.78     & 6.29  & 1.19  & 0.61     & 5.60     & 0.84  \\
bel\_Cyrl & Low  & 1.78  & 1.71    & 2.00     & 2.99  & 1.51  & 1.34     & 1.39     & 1.63  \\
ben\_Beng & High & 0.16  & 1.52    & 2.43     & 1.59  & 3.56  & 0.76     & 1.53     & 12.10 \\
\bottomrule
\end{tabular}}
\caption{\label{tab:und_flores_to_1}
\textbf{Understanding Capability (Part I):} BLEU scores on the FLORES benchmark~\cite{goyal-etal-2022-flores} and average scores on low-resource (Avg\_L) and high-resource (Avg\_H) languages. We report on translating from the other languages into English.
}
\end{table*}

\begin{table*}[t]
\centering
\resizebox{\textwidth}{!}{
\begin{tabular}{ll|rrrr|rrrr}
\toprule
\#Langs.             & Type   & LLaMA & \bxllama & \sftllama & \md    & BLOOM & \bxbloom & \sftbloom & \md    \\
\midrule
luo\_Latn & Low  & 3.14  & 0.78    & 1.96     & 3.40  & 1.32  & 1.94     & 8.60     & 6.47  \\
lvs\_Latn & High & 1.65  & 1.62    & 2.44     & 2.37  & 1.34  & 1.13     & 1.74     & 1.98  \\
mal\_Mlym & Low  & 1.18  & 0.92    & 0.57     & 1.46  & 2.94  & 0.92     & 2.98     & 4.08  \\
mar\_Deva & Low  & 2.06  & 1.75    & 2.60     & 1.66  & 2.64  & 1.38     & 1.78     & 7.09  \\
mkd\_Cyrl & High & 11.35 & 5.74    & 8.93     & 12.52 & 2.24  & 1.21     & 1.46     & 4.67  \\
mlt\_Latn & High & 1.61  & 2.67    & 2.56     & 3.96  & 2.05  & 1.65     & 2.72     & 2.10  \\
khk\_Cyrl & Low  & 1.46  & 1.00    & 1.60     & 1.46  & 1.31  & 1.40     & 1.49     & 1.62  \\
\bottomrule
\end{tabular}}
\caption{\label{tab:und_flores_to_2}
\textbf{Understanding Capability (Part II):} BLEU scores on the FLORES benchmark~\cite{goyal-etal-2022-flores} and average scores on low-resource (Avg\_L) and high-resource (Avg\_H) languages. We report on translating from the other languages into English.
}
\end{table*}

\end{document}